\documentclass{rtp_tech_report}

\usepackage{amsmath,amsfonts,bm}

\def\eqref#1{equation~\ref{#1}}

\def\1{\bm{1}}

\DeclareMathAlphabet{\mathsfit}{\encodingdefault}{\sfdefault}{m}{sl}
\SetMathAlphabet{\mathsfit}{bold}{\encodingdefault}{\sfdefault}{bx}{n}

\DeclareMathOperator*{\argmax}{arg\,max}

\usepackage{amsmath,amssymb}
\usepackage{algorithm}
\usepackage{algorithmic}
\usepackage{booktabs}
\usepackage{array}
\usepackage{graphicx}
\usepackage[table]{xcolor}
\usepackage{tikz}
\usetikzlibrary{arrows.meta,calc,decorations.pathreplacing,positioning}
\usepackage{pgfplots}
\pgfplotsset{compat=1.18}
\usepgfplotslibrary{groupplots,fillbetween}
\usepackage{hyperref}
\hypersetup{
  pdftitle={Adversarial Gradient Forcing for Robust Few-Step Autoregressive Video Generation},
  pdfauthor={Chenxi Xia, Yixuan Xu, Shan Liu, Jiapeng Wang, Maohua Li, Hanlin Tang, Kan Liu, Tao Lan, Weiyi Lu, Xinggao Liu}
}
\usepackage{url}
\usepackage{placeins}
\usepackage{wrapfig}
\usepackage{needspace}

\newcommand{\stopgrad}{\operatorname{sg}}
\newcommand{\RMS}{\operatorname{RMS}}
\newcommand{\inner}[2]{\left\langle #1,#2\right\rangle}

\definecolor{agfBestCell}{HTML}{F4E4DA}
\definecolor{agfSecondCell}{HTML}{E4EDF7}
\newcommand{\bestcell}[1]{\cellcolor{agfBestCell}{\bfseries #1}}
\newcommand{\secondcell}[1]{\cellcolor{agfSecondCell}{#1}}

\title{Adversarial Gradient Forcing for Robust Few-Step Autoregressive Video Generation}

\author[1,2,\sharp]{\mbox{Chenxi Xia}}
\author[2]{\mbox{Yixuan Xu}}
\author[2]{\mbox{Shan Liu}}
\author[2]{\mbox{Jiapeng Wang}}
\author[2,3]{\mbox{Maohua Li}}
\author[2]{\mbox{Hanlin Tang}}
\author[2]{\mbox{Kan Liu}}
\author[2]{\mbox{Tao Lan}}
\author[2,\dagger]{\mbox{Weiyi Lu}}
\author[1,\S]{\mbox{Xinggao Liu}}

\affiliation[1]{Zhejiang University}
\affiliation[2]{Alibaba Group}
\affiliation[3]{Nanjing University}

\contribution[\S]{Corresponding author}
\contribution[\dagger]{Project lead}
\contribution[\sharp]{Work done during internship at Alibaba}

\checkdata[E-mail]{\texttt{\{chenxi.x, lxg\}@zju.edu.cn, weiyi.lwy@alibaba-inc.com}}
\checkdata[Project page]{\url{https://wuli-art.github.io/Adversarial-Gradient-Forcing/}}
\checkdata[Code]{\url{https://github.com/wuli-art/Adversarial-Gradient-Forcing}}

\abstract{
Autoregressive video generation models generate videos sequentially, with each new block conditioned on previously generated blocks. By training on the model's own rollouts, Self Forcing and its follow-up methods substantially mitigate exposure bias caused by the mismatch between training- and inference-time history distributions. However, these models still suffer from error accumulation during long-horizon generation. We identify a key underlying limitation:
the model remains highly sensitive to errors in its generated history, allowing small deviations to be amplified through subsequent generations.
We refer to this underexplored problem as the local history robustness gap.
To address this gap, we introduce Adversarial Gradient Forcing (AGF), a framework designed to improve the robustness of few-step autoregressive video models to perturbations and accumulated errors in their own generated histories. AGF uses distribution matching gradients from future block predictions to select adversarial perturbations of self-generated histories. It then jointly trains the generator on the original and perturbed histories to improve the robustness of future generation.
Extensive experiments show that AGF achieves the highest VBench Total score on 60-second video generation and the best visual quality on four-minute video generation. Notably, AGF achieves such long-horizon robustness while requiring only 5-second rollouts for training.

}

\begin{document}

\maketitle

\section{Introduction}
\label{sec:introduction}
Video generation has progressed from early diffusion-based generative models~\citep{ho2022videodiffusion} to large-scale systems capable of synthesizing high-quality, temporally coherent videos~\citep{blattmann2023stablevideo,chen2024videocrafter2,yang2024cogvideox,kong2024hunyuanvideo,wan2025wan}. Autoregressive approaches further extend video generation by producing successive video blocks conditioned on previously generated blocks~\citep{henschel2025streamingt2v,longcat2025video,gao2024ca2vdm}. Recent methods further combine autogressive modeling with few-step denoising to reduce inference latency~\citep{yin2025causalvideo,lin2025aapt,zhu2026causalforcing}, enabling streaming and interactive video generation~\citep{yang2026longlive,shin2026motionstream,savva2026solaris}. Among these methods, self forcing further narrows the training--inference gap by optimizing on self-generated rollouts with Distribution Matching Distillation (DMD)~\citep{huang2025selfforcing,yin2024dmd,yin2024dmd2}, and subsequent methods extend this paradigm through longer training trajectories and additional supervision~\citep{cui2025selfforcingpp,yu2026videomirai,zhuang2026sgf,liu2026opsdv}. Together, these methods aim to make autoregressive video generation more efficient and reliable over long horizons.

However, current autoregressive video generation models still suffer from the notorious error accumulation problem during long-horizon generation~\citep{yang2026longlive,liu2026opsdv}.
Since each generated block becomes part of the conditioning history, errors in earlier blocks can propagate through subsequent generation and progressively degrade visual quality~\citep{huang2025selfforcing,yang2026longlive,li2025svi}.
We identify a key cause of error accumulation: the model remains highly sensitive to errors in its self-generated history.
In Section~\ref{sec:gap-formulation}, we present details on the error sensitivity and refer to it as the \emph{local history robustness gap}. We further show that improving robustness to these errors substantially mitigates error accumulation and stabilize long-horizon generation.

\begin{figure}[!t]
  \centering
  \setlength{\abovecaptionskip}{4pt}
  \setlength{\belowcaptionskip}{0pt}
  \input{figure/source/intro_comparison.tex}
  \caption{\textbf{An example of four-minute video generation.} Self Forcing and AGF generate four-minute videos using the same ODE initialization. With Self Forcing, visual artifacts emerge and progressively worsen over time, while AGF maintains coherent subjects and scene structure.}
  \label{fig:intro-case022}
\end{figure}

To address this gap, we introduce \emph{Adversarial Gradient Forcing} (AGF), a novel training framework for improving long-horizon visual quality in few-step autoregressive video generation. Rather than merely training on the model's own rollouts, AGF deliberately constructs challenging histories according to how strongly they impair the subsequent generation. By explicitly exposing the model to such challenging histories during training, AGF improves the robustness against accumulated errors, and suppresses their propagation into subsequent blocks. More specifically, AGF backpropagates the DMD gradient from future rollout blocks to a randomly selected history block, and uses the resulting gradient to construct a bounded adversarial perturbation to the selected history block.


Given the adversarial perturbation, AGF then recomputes block predictions under both the original and perturbed histories and jointly optimizes the generator with DMD. The original-history branch preserves performance under the model's standard generation conditions, while the perturbed-history branch improves robustness to errors in the generated history. DMD is evaluated independently on the rollout produced by each branch, yielding the training gradients for joint optimization.

We further introduce a \emph{prediction consistency regularizer} during early training to prevent adversarial perturbations from driving the generated trajectory too far from the local data manifold. This regularization term further improves consistency in viewpoint and scene structure.


We conduct extensive experiments to evaluate AGF. Despite being trained only on five-second rollouts, AGF achieves the highest 60-second VBench Total score and the best four-minute visual quality among all five evaluated baselines, while remaining competitive on five-second generation. More importantly, AGF sustains stable visual quality throughout four-minute generation, whereas all baselines progressively degrade as generation proceeds (Section~\ref{sec:temporal-decay}), as illustrated in Figure~\ref{fig:intro-case022}.

Our main contributions are summarized as follow:
\begingroup
\setlength{\leftmargini}{1.5em}
\setlength{\labelwidth}{1em}
\setlength{\labelsep}{0.5em}
\begin{itemize}
  \item We analyze how small errors in generated histories propagate to subsequent generation and identify the local history robustness gap in existing methods.
  \item We propose AGF, which selects adversarial history perturbations using gradients from future and incorporates them into on-policy distribution matching to improve robustness to history errors.
  \item Extensive experiments demonstrate that AGF improves visual quality
  and reduces degradation during long video generation. Ablation studies
  validate the effectiveness of each component.
\end{itemize}
\endgroup

\Needspace{8\baselineskip}
\section{Related Work}
\label{sec:related-work}

\subsection{Training Few-Step Autoregressive Video Models}

Early approaches differ primarily in how they construct the histories used for autoregressive generation. Teacher Forcing conditions on ground-truth histories~\citep{bengio2015scheduled,lamb2016professor}, whereas Diffusion Forcing relaxes this dependency by independently corrupting video frames with noise~\citep{chen2024diffusionforcing}. CausVid then combines Diffusion Forcing with Distribution Matching Distillation (DMD) to enable few-step autogressive generation~\citep{yin2025causalvideo,yin2024dmd,yin2024dmd2}. Self Forcing takes a novel direction by training the model on its own autoregressive rollouts, thereby reducing the mismatch between training- and inference-time history distributions~\citep{huang2025selfforcing}. AAPT further introduces discriminator-based supervision over generated histories~\citep{lin2025aapt}. Subsequent methods built on this paradigm propose improved causal initialization~\citep{zhu2026causalforcing,zhao2026causalforcingpp,zheng2026causalrcm}, longer self-generated training trajectories~\citep{yang2026longlive,cui2025selfforcingpp,chen2026contextforcing,luo2026shotstream}, joint denoising across adjacent blocks~\citep{liu2025rollingforcing,li2026msforcing}, and gradient propagation through recomputed history representations~\citep{savva2026solaris,zhuang2026sgf}. Despite these improvements, robustness against errors already present in generated histories remains underexplored. The proposed AGF targets this problem by improving the model's tolerance to imperfect self-generated histories, thereby mitigating error accumulation over long-horizon generation while requiring only short training rollouts.

\subsection{Training with Imperfect Histories}

Recent studies have begun to investigate autoregressive video generation under imperfect history conditions. GameNGen injects noise into conditioning frames, while Helios applies predefined frame corruptions to improve long-horizon video generation~\citep{valevski2024gamengen,yuan2026helios}. Resampling Forcing constructs imperfect histories by self-resampling real videos, Stable Video Infinity reuses prediction errors from previous generations, and BAgger builds training trajectories from progressively drifting rollouts~\citep{guo2025selfresampling,li2025svi,po2025bagger}. Salt further diversifies history quality through cache perturbations and feature alignment, whereas OPSD-V leverages real-video context to provide teacher supervision~\citep{ge2026salt,liu2026opsdv}. Despite these efforts, none of these methods explicitly assess how individual history perturbations affect downstream generations and use this impact to select informative training histories.

Adversarial training improves robustness by constructing input perturbations that increase the training loss~\citep{szegedy2014intriguing,goodfellow2015adversarial,madry2018adversarial}. BadWorld studies adversarial perturbations to the conditioning context of fixed world models~\citep{shen2026badworld}.
In contrast, AGF introduces adversarial history selection into on-policy distribution matching, using the perturbed histories to train the model for improved robustness.


\begin{figure}[!htbp]
  \centering
  \includegraphics[width=\linewidth]{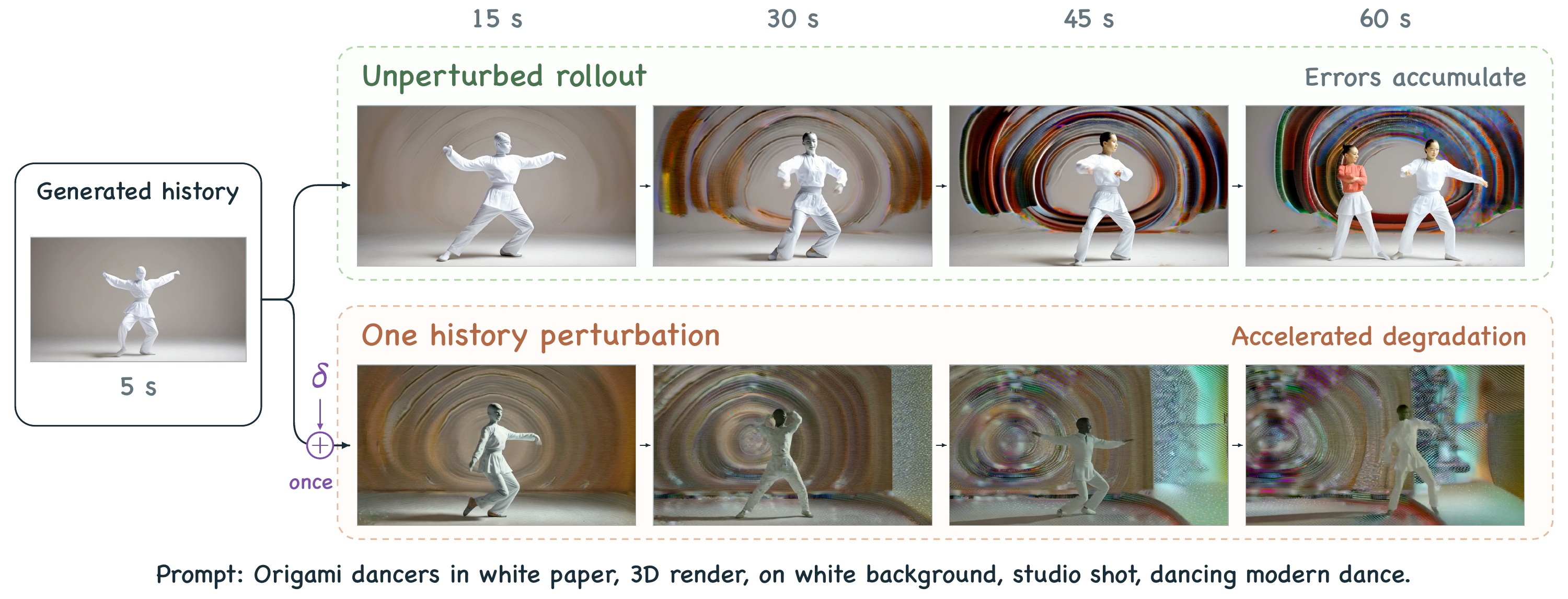}
  \caption{\textbf{A single perturbation leads to severe visual degradation in long-horizon autogressive video generation.} The top row shows an unperturbed Self-Forcing rollout, while the bottom row uses the same model and identical sampling randomness for future generation, differing only by a one-time perturbation applied to the generated history. Compared with the unperturbed rollout, concentric background artifacts emerge earlier and progressively intensify in the perturbed rollout, accompanied by increasingly severe subject distortion.}
  \label{fig:local-history-gap}
\end{figure}

\begingroup
\raggedbottom
\section{Method}
\label{sec:method}

In this section, we first introduce the local history robustness gap, where small errors in generated histories can lead to catastrophic visual degradation in subsequent generation. Building on this observation, we then present AGF, including its adversarial perturbation selection strategy and training pipeline. We further introduce a regularization term that encourages the adversarial perturbations to remain close to the generator's data manifold. Finally, we visualize the learned adversarial perturbations and analyze their effects on both generated histories and subsequent generation.

\subsection{The Local History Robustness Gap}
\label{sec:gap-formulation}
Autoregressive video generation models condition each new block on previously generated ones, making generation quality highly sensitive to errors accumulated in the history. Even a small perturbation to a generated history block can propagate forward and substantially degrade subsequent generation. As illustrated in Figure~\ref{fig:local-history-gap}, perturbing a single history block causes visual artifacts to emerge earlier and become progressively more severe compared with the unperturbed rollout. We refer to the model’s vulnerability to small history errors that degrade future generation as the \emph{local history robustness gap}.



Motivated by this observation, AGF aims to identify perturbations on the generated history that maximally impair subsequent generation, and trains the model to produce high-quality continuations from both clean and perturbed histories. In Appendix~\ref{app:causal-error-propagation}, we provide a theoretical analysis of the error propagation mechanism, showing that perturbations introduced into the history and errors accumulated during autoregressive generation share the same underlying mechanism.


\subsection{Adversarial Gradient Forcing}
\label{sec:history-direction}
Rather than selecting the perturbation that most changes future predictions, AGF selects the one that most increases the DMD loss on future blocks. High sensitivity to history does not by itself imply a distribution-matching error: a generator can respond strongly to its history while its continuations remain correctly distributed (Appendix~\ref{app:distribution-response}).


DMD trains a few-step autoregressive video generator, parameterized by $\theta$, to match a teacher-defined target distribution by minimizing the reverse KL divergence. At noise level $\tau$, the frozen teacher estimates the score $s_{\mathrm{real}}$ of the noisy target distribution, while a fake-score model, trained online on samples from the generator, estimates the score $s_{\mathrm{fake}}$ of the noisy generator-induced distribution. Given a generated video $X_0=(x_0^{(1)},\ldots,x_0^{(K)})$ represented by $K$ clean latent blocks, DMD perturbs it with Gaussian noise to obtain $X_\tau$, evaluates both scores at $X_\tau$, and uses their difference to construct the distribution-matching gradient $g_{\mathrm{DM}}(X_0)$:

\begin{equation}
  g_{\mathrm{DM}}(X_0)
  \propto s_{\mathrm{fake}}(X_\tau,\tau)-s_{\mathrm{real}}(X_\tau,\tau)
  \propto \widehat X_{0,\mathrm{fake}(X_\tau,\tau)}-\widehat X_{0,\mathrm{real}}(X_\tau,\tau).
  \label{eq:dmd-gradient}
\end{equation}
Here, $\widehat X_{0,\mathrm{real}}$ and $\widehat X_{0,\mathrm{fake}}$
are the two models' predictions.
DMD backpropagates $g_{\mathrm{DM}}(X_0)$ through the generator
to update $\theta$ by gradient descent.
Further details are available in
Appendix~\ref{app:gradient-carrier}.


\paragraph{(a) Sample autoregressive rollout trajectories.}
To supervise different denoising stages, AGF first samples a generator
timestep $t^\star$ for each rollout and uses the same timestep across all $K$ blocks.  It then performs an autoregressive rollout without retaining the computation graph. For each block $k$, the generator takes the noisy latent block $\bar x_{t^\star}^{(k)}$ as input and predicts the corresponding clean latent block $\bar x_0^{(k)}$.
Rollout trajectories are recorded as $\bar X_{t^\star}=(\bar x_{t^\star}^{(1)},\ldots,\bar x_{t^\star}^{(K)})$ and $\bar X_0=(\bar x_0^{(1)},\ldots,\bar x_0^{(K)})$ for use in the following stages. AGF then randomly selects a block $j<K$ to perturb.

\FloatBarrier
\begin{figure}[!t]
  \centering
  \includegraphics[width=\linewidth]{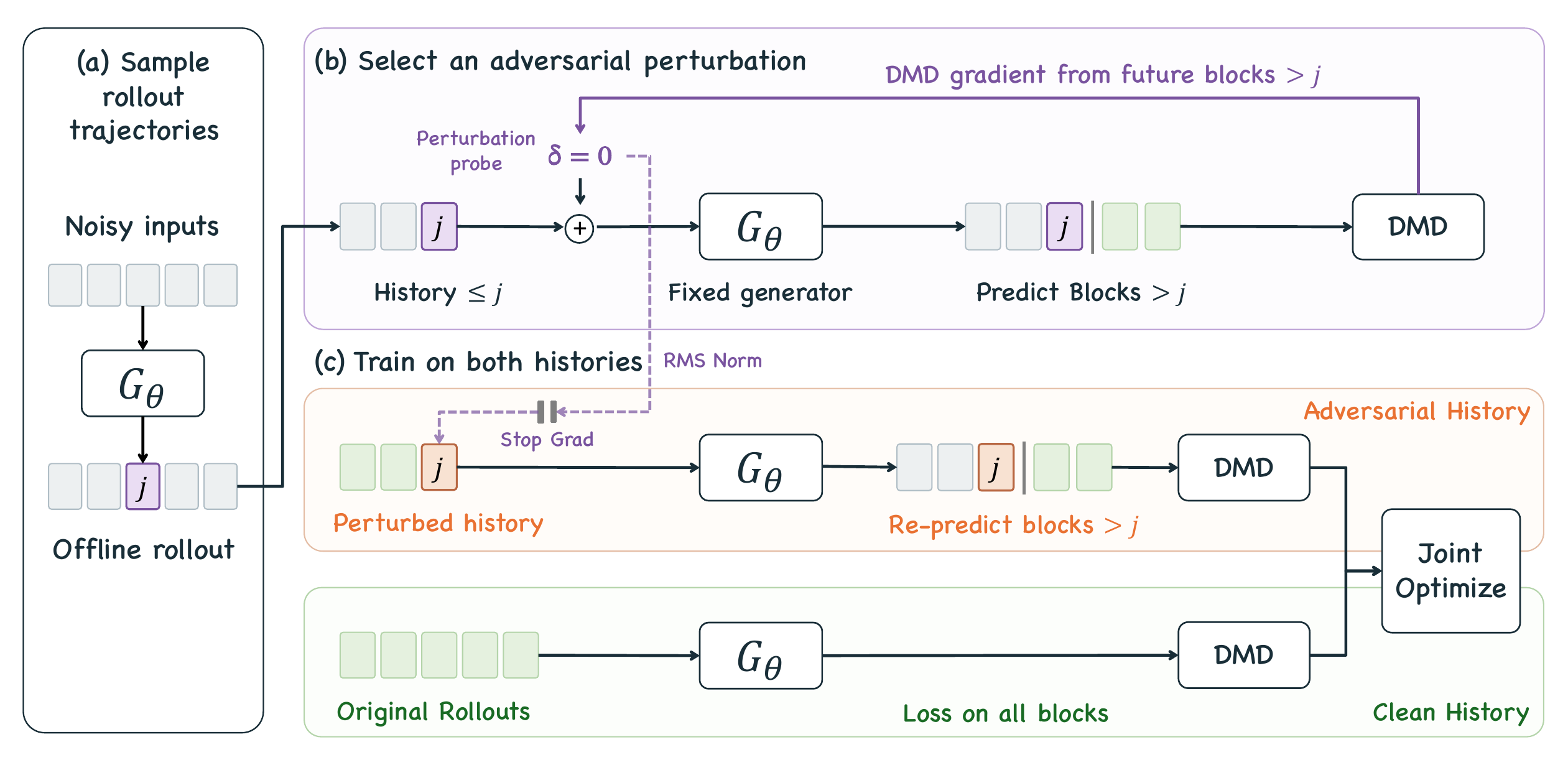}
  \caption{\textbf{Adversarial Gradient Forcing.}
    (a) Sample autoregressive rollout trajectories offline without gradient.
    (b) Use DMD gradients from blocks after $j$ to select a history perturbation that most adversely affects subsequent generations.
    (c) With the perturbation fixed, train jointly on the clean and perturbed histories. $G_\theta$ denotes the generator in DMD parameterized by $\theta$.}
  \label{fig:agf-overview}
\end{figure}

\paragraph{(b) Select an adversarial history perturbation.}
To find a small perturbation to the selected block $j$ that most adversely
affects subsequent generation, AGF then adds a zero-initialized perturbation $\delta$
on the clean latent block $\bar x_0^{(j)}$ to probe how changes to this block affect subsequent generation.

Given the offline rollout trajectory $\bar X_{t^\star}$, we recompute the clean latent predictions in parallel for all blocks $k>j$ with computation graph enabled. Specifically, the generator first encodes the recorded clean latent blocks into key--value (K/V) representations, with the perturbation $\delta$ added to $\bar x_0^{(j)}$. Then, for each block $k>j$, the generator takes the recorded noisy latent $\bar x_{t^\star}^{(k)}$ as input and conditions on the K/V representations constructed from the preceding latent blocks, where block $j$ is perturbed by $\delta$. We denote the resulting predictions for all blocks after $j$ by $\widehat Y_j(\delta)$. This forward pass preserves a differentiable path from $\widehat Y_j(\delta)$ back to $\delta$ (Appendix~\ref{app:gradient-boundaries}).

Next, we compute the distribution--matching gradient on the full rollouts, and backpropagate only its entries $\bar g_j$ from blocks after $j$ to $\delta$, obtaining the history gradient $u_j$, along which
the DMD loss increases most rapidly.
This backpropagation gives $u_j=J_j^\top\bar g_j$, where $J_j$ is the
Jacobian of the later predictions $\widehat Y_j(\delta)$ with respect
to $\delta$, evaluated at $\delta=0$
(Appendix~\ref{app:tangent-pullback}).

\Needspace{8\baselineskip}
The history gradient $u_j$ may differ across the latent frames in block $j$. To obtain a shared direction $\widetilde u_j$, we average $u_j$ across the latent frame dimension and use the resulting mean as the perturbation
direction for every frame in block $j$.
This shared direction $\widetilde u_j$ is then rescaled to a root-mean-square (RMS)
magnitude of $\epsilon$, giving
$\delta_j^\star=\stopgrad(\epsilon\,\widetilde u_j/\RMS(\widetilde u_j))$,
where $\epsilon$ controls the perturbation strength.
The resulting $\delta_j^\star$ is the adversarial history perturbation
selected for training (Appendix~\ref{app:pullback}).
The operator $\operatorname{sg}$ stops gradients through perturbations, keeping $\delta_j^\star$ fixed during generator training.

\paragraph{(c) Train on both histories.}
After adding the selected adversarial perturbation $\delta_j^\star$ to the recorded
clean latent block $\bar x_0^{(j)}$, we then recompute denoise predictions from $\bar X_{t^\star}$ at $t^\star$ under both the clean and perturbed histories.
The clean and adversarial branches use the original and perturbed histories
to produce full video predictions $X_0^{\mathrm{clean}}$ and
$X_0^{\mathrm{adv}}$, respectively.

We evaluate DMD on each full rollout to obtain the corresponding gradients $g^{\mathrm{clean}}$ and $g^{\mathrm{adv}}$. The surrogate loss $\ell_{\mathrm{DM}}(X_0;g)$ applies the fixed DMD gradient $g$ to the prediction $X_0$ (Appendix~\ref{app:gradient-carrier}). We train the generator with the joint learning objective:

\begin{equation}
  \mathcal L_{\mathrm{AGF}}
  =\ell_{\mathrm{DM}}(X_0^{\mathrm{clean}};g^{\mathrm{clean}})
  +\ell_{\mathrm{DM}}^{>j}(X_0^{\mathrm{adv}};g^{\mathrm{adv}}).
  \label{eq:agf-objective}
\end{equation}

The clean loss branch $\ell_{\mathrm{DM}}$ averages over all blocks within one rollout, while the adversarial loss branch $\ell_{\mathrm{DM}}^{>j}$ only averages over blocks after the perturbed block $j$, focusing training on predictions conditioned on the perturbed history. The fake-score model in DMD is also trained on a weighted mixture of samples from both branches (Appendix~\ref{app:fake-score}). Algorithm~\ref{alg:agf} in Appendix~\ref{app:agf-algorithm} summarizes the complete training procedure.


\subsection{Prediction Consistency Regularizer}
\label{sec:prediction-consistency}


To further improve consistency between the clean and adversarial branches, we introduce a \emph{prediction consistency regularizer} that aligns velocity predictions of the generator at the first denoising step. Enforcing consistency at this initial stage encourages the two branches to preserve similar global scene structures, while allowing subsequent denoising steps to independently refine local appearance and fine-grained details.


Let $v_{>j}^{\mathrm{clean}}$ and $v_{>j}^{\mathrm{adv}}$ denote generator's predicted velocity for all blocks after $j$ at the first denoising step. The regularization loss is written as:
\begin{equation}
  \mathcal L_{\mathrm{reg}}
  =\operatorname{MSE}\!\left(
    v_{>j}^{\mathrm{adv}},\stopgrad(v_{>j}^{\mathrm{clean}})
  \right).
  \label{eq:prediction-consistency}
\end{equation}
$\mathcal L_{\mathrm{reg}}$ averages over all blocks after $j$ with the prediction from the clean branch detached, such that the regularizer only affects the generator through the adversarial branch.

During the early training stage, we add this regularizer to the AGF learning objective
to encourage consistency:
$\mathcal L_{\mathrm{train}}=\mathcal L_{\mathrm{AGF}}+
\lambda_s\mathcal L_{\mathrm{reg}}$, where $\lambda_s$ is the regularization weight at training iteration $s$.
We then remove the regularizer and continue optimizing
$\mathcal L_{\mathrm{AGF}}$ alone to further improve visual quality.
Appendix~\ref{app:prediction-consistency-details} provides details on the regularization weight scheduler. Section~\ref{sec:ablations} conducts ablation studies on different choices for the regularizer.

\subsection{Visualizing the Effects of History Perturbations in AGF}
\label{sec:gradient-placement}

The perturbations selected by AGF tend to target vulnerable scene features in pixel space, including fine textures, subtle color variations, and complex structures. By training on these perturbed histories, AGF directs the model's attention toward fine-grained details and complex structures whose corruption has a particularly detrimental effect on subsequent generation, thereby improving the generator's ability to preserve these features over long rollouts.


Figure~\ref{fig:gradient-phenotype} compares the effects of AGF-selected perturbations with those of Gaussian noise in pixel space, with both applied to the same history block. As shown, AGF perturbations selectively affect vulnerable scene details, particularly fine hull textures, rigging around the masts, and sail boundaries. Moreover, the second-to-last subfigure shows that the influence of perturbations from AGF remains pronounced in later generated frames, indicating their persistent effect on subsequent generation. In contrast, Gaussian noise induces more spatially uniform changes in the history block and results in substantially weaker deviations in later frames. In addition to this example, Appendix~\ref{app:gradient-phenotype-gallery} provides a more comprehensive analysis of this behavior using additional cases.



\begin{figure}[H]
  \centering
  \includegraphics[width=\linewidth]{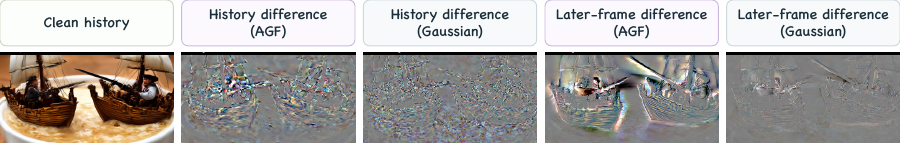}
  \caption{\textbf{Effects of selected and Gaussian history perturbations.}
  The leftmost clean history serves as the scene reference.
  \emph{History difference} shows the pixel difference between corresponding
  frames decoded from the perturbed and original histories.
  \emph{Later-frame difference} shows the pixel difference between
  corresponding later frames generated from two separate histories.}
  \label{fig:gradient-phenotype}
\end{figure}

\Needspace{5\baselineskip}
\section{Experiments}
\label{sec:experiments}

\subsection{Experimental Setup}
\label{sec:exp-setup}

\noindent\textbf{Training.}
We train a four-step causal video generator based on Wan2.1-T2V-1.3B~\citep{wan2025wan}, initialized from the ODE initialization checkpoint released by Self-Forcing~\citep{huang2025selfforcing}.
We use Wan2.1-T2V-14B as the frozen teacher and train on
the filtered and LLM-extended VidProM~\citep{wang2024vidprom} text prompts same as Self-Forcing. Training uses five-second rollouts with an effective batch size of 64, consisting of 200 iterations of AGF with prediction consistency regularization followed by 200 iterations without regularization. All experiments are conducted on eight NVIDIA H20 GPUs.
Full training details are available in Appendix~\ref{app:experiment-details}.

\noindent\textbf{Evaluation.}
16 VBench dimensions~\citep{huang2023vbench}
on the full 944-prompt suite at 5 seconds, and a randomly sampled
240-prompt subset at 60 seconds are evaluated. At 240 seconds, we further use 120 prompts randomly sampled from MovieGenVideoBench\_extended
~\citep{polyak2024moviegen}.

We report VBench Total at 5 seconds and Quality, Semantic, and
Total at 60 seconds. At 240 seconds, we evaluate the six VBench-Long
dimensions supported for custom prompts~\citep{huang2025vbenchpp}.
For visual quality, we follow LongLive~\citep{yang2026longlive} and use five VBench-Long dimensions: subject consistency, background consistency, motion smoothness, aesthetic quality, and imaging quality.
We define \textbf{Five-metric Quality (Q5)} as their average results and use it as our primary quality score.
Full protocols and checkpoint details appear in
Appendix~\ref{app:experiment-details}.

\noindent\textbf{Baselines.}
We compare against Self-Forcing~\citep{huang2025selfforcing},
Causal Forcing~\citep{zhu2026causalforcing},
LongLive~\citep{yang2026longlive},
Self Gradient Forcing (SGF)~\citep{zhuang2026sgf},
and OPSD-V~\citep{liu2026opsdv}.
For OPSD-V, we evaluate the official checkpoint post-trained from Self-Forcing.
We reproduce SGF from the same ODE initialization as AGF, using SGF's
official training settings.
All methods adopt the same inference settings, with 3 sink and 9 rolling latent frames.

\begin{table}[!htbp]
  \centering
  \caption{\textbf{Quantitative comparison with relevant baselines.}
    We benchmark all methods from native 5-second generation to 240-second long-horizon generation.
    Except for Dynamic Degree, orange and blue indicate the best and second-best results, respectively; the best results are bolded. Higher Dynamic Degree may also arise from severe camera drift or visual corruption and does not necessarily indicate better generation quality.}
  \label{tab:long-results}
  {\scriptsize
  \setlength{\tabcolsep}{2.1pt}
  \renewcommand{\arraystretch}{1.05}
  \begin{tabular*}{\textwidth}{@{\extracolsep{\fill}}l
    *{3}{w{c}{25pt}} w{c}{36pt}
    *{6}{w{c}{25pt}} w{c}{36pt}@{\hspace{\tabcolsep}}}
    \toprule
    & \multicolumn{1}{c}{5 s}
    & \multicolumn{3}{c}{60 s}
    & \multicolumn{7}{c}{240 s} \\
    \cmidrule(lr){2-2}\cmidrule(lr){3-5}\cmidrule(lr){6-12}
    Method & Total$\uparrow$
    & Quality$\uparrow$ & Semantic$\uparrow$ & Total$\uparrow$
    & Subj.$\uparrow$ & Back.$\uparrow$ & Motion$\uparrow$
    & Aesth.$\uparrow$ & Imaging$\uparrow$ & Dyn.$^{\dagger}$
    & \textbf{Q5}$\uparrow$ \\
    \midrule
    Self-Forcing & \secondcell{82.21} & 82.95 & 66.01 & 79.56 & 96.36 & 95.79 & 97.95 & 54.97 & 67.26 & 63.19 & 82.47 \\
    Causal Forcing & \bestcell{82.94} & \secondcell{83.82} & 63.88 & 79.83 & 94.44 & 94.82 & 95.99 & 55.29 & 68.41 & 91.41 & 81.79 \\
    LongLive & 80.34 & 82.11 & \bestcell{70.59} & 79.81 & 97.08 & \secondcell{96.24} & \bestcell{98.60} & 58.29 & 67.86 & 49.53 & 83.62 \\
    SGF & 82.08 & 83.56 & 64.09 & 79.67 & \secondcell{97.11} & 96.10 & 98.18 & \secondcell{58.32} & \secondcell{70.63} & 51.08 & \secondcell{84.07} \\
    OPSD-V & 81.95 & 83.16 & 67.67 & \secondcell{80.06} & 96.29 & 95.64 & 97.91 & 56.60 & 67.77 & 65.10 & 82.84 \\
    \midrule
    AGF (ours) & 82.03 & \bestcell{84.67} & \secondcell{68.66} & \bestcell{81.47} & \bestcell{97.32} & \bestcell{96.38} & \secondcell{98.31} & \bestcell{61.20} & \bestcell{70.91} & 62.54 & \bestcell{84.82} \\
    \bottomrule
  \end{tabular*}}
\end{table}

\subsection{Main Long-Horizon Benchmark}
\label{sec:long-results}

Table~\ref{tab:long-results} shows that AGF maintains comparable short-horizon performance while achieving state-of-the-art results at longer horizons.
At five seconds, AGF performs comparably to Self-Forcing, indicating that adversarial training does not compromise short-horizon generation quality.
By 60 seconds, AGF already outperforms the baselines, achieving the best Quality and Total scores among all methods.
This advantage becomes more pronounced at 240 seconds: AGF ranks first in four of the five quality dimensions and achieves the highest Q5 score of 84.82. Complete per-dimension results are provided in Appendix~\ref{app:complete-benchmark}.
These results indicate that AGF primarily improves the generator's ability to preserve visual quality as autoregressive errors accumulate over long rollouts.




Figure~\ref{fig:long-horizon-cases} presents a representative long-horizon comparison on a coastal church scene.
As generation proceeds, the competing methods variously suffer from subject loss, compositional drift, or structured artifacts, whereas AGF preserves both the church and coastline through the final frame.
Appendix~\ref{app:full-case-comparisons} provides additional comparisons on more diverse scenes.

Additional visual comparisons reveal failure modes not fully captured by benchmark scores.
In Causal Forcing and several other methods, high Dynamic Degree often coincides with camera drift or accumulating artifacts, showing that stronger motion does not ensure better long-horizon quality (Appendix~\ref{app:dynamic-degree}).
Meanwhile, AGF achieves high Dynamic Degree in genuinely dynamic scenes, such as clay sculpting, salad tossing, and character movement, while preserving the subjects and surrounding scenes (Appendix~\ref{app:high-dynamic-quality}).
For Q5, in some scenes, methods with Q5 scores comparable to or higher than AGF's still exhibit subject duplication or background drift (Appendix~\ref{app:similar-quality}).



\begin{figure}[!t]
  \centering
  \input{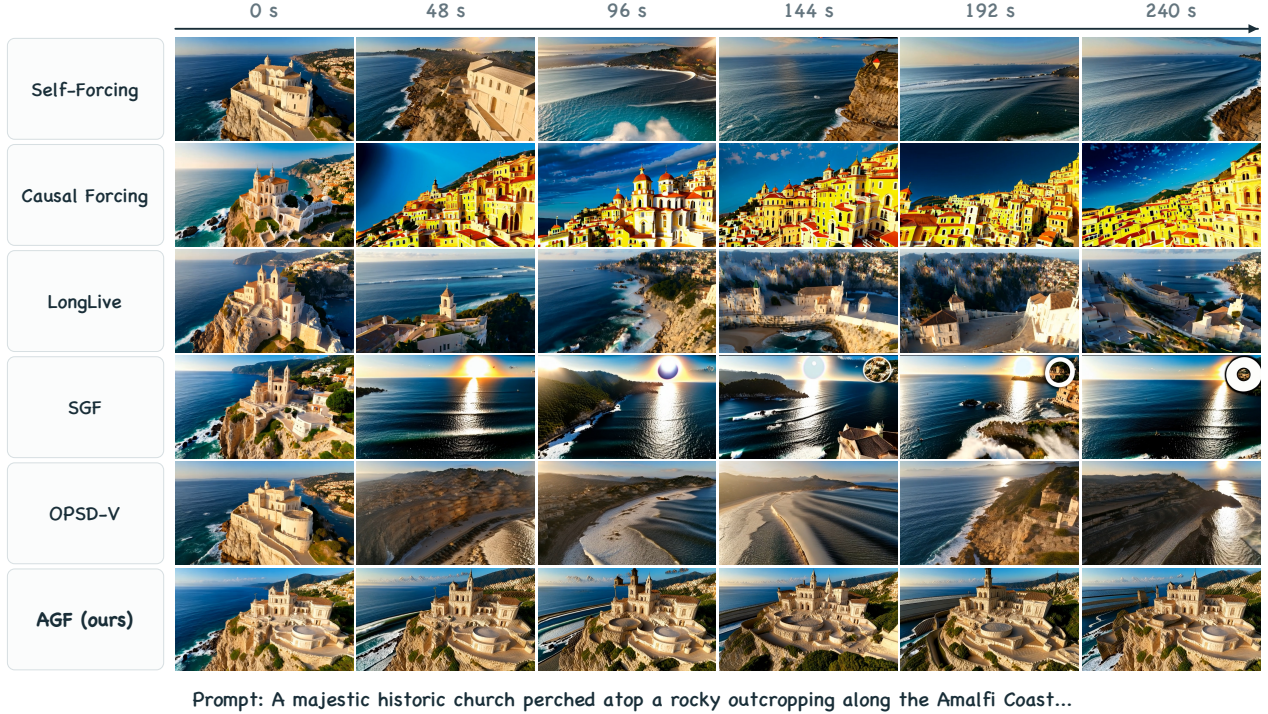}
  \caption{\textbf{Long-horizon qualitative comparison.} Six-method rollouts
on the coastal church scene at 0, 48, 96, 144, 192, and 240 seconds.
AGF better preserves the subject and scene composition.}
  \label{fig:long-horizon-cases}
\end{figure}

\subsection{Stability over Rollout Time}
\label{sec:temporal-decay}

We next evaluate the stability of visual quality throughout four-minute generation.
Building on the evaluation in Section~\ref{sec:long-results}, we track Q5 over consecutive 20-second intervals and report its change relative to the first interval. Figure~\ref{fig:temporal-decay} shows that AGF maintains stable visual quality throughout the rollout, with a first-to-last change of $+0.03$ pp, whereas all five baselines decline by $0.96$--$2.61$ pp. Detailed results for all methods are provided in Appendix~\ref{app:temporal-unperturbed}, while Appendix~\ref{app:temporal-perturbed} further demonstrates AGF's stronger robustness under two temporary history perturbations.



\begin{figure}[!t]
  \begin{minipage}[t]{0.605\linewidth}
    \vspace{0pt}
    \includegraphics[width=\linewidth,height=140pt,keepaspectratio]{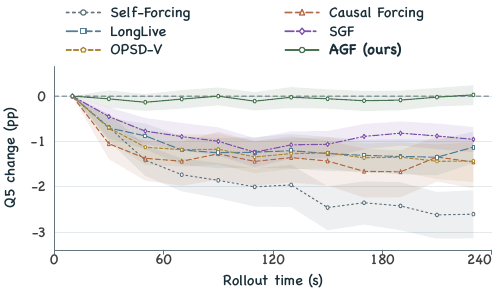}\par
    \caption{\textbf{Visual quality over time.} Local Q5 change from the
    first 20 seconds, averaged over 120 prompts; shading shows pointwise
    95\% bootstrap confidence intervals.}
    \label{fig:temporal-decay}
  \end{minipage}\hfill
  \begin{minipage}[t]{0.375\linewidth}
    \vspace{0pt}
    \includegraphics[width=\linewidth,height=140pt,keepaspectratio]{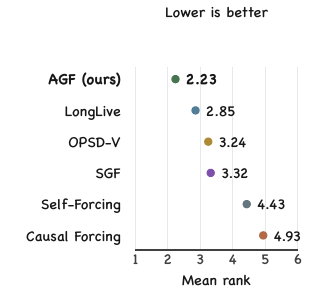}\par
    \caption{\textbf{Human ranking.} Mean ranks from 240 tasks covering
    120 prompts (lower is better).}
    \label{fig:ranking-mean}
  \end{minipage}
\end{figure}

\Needspace{95pt}
\subsection{Human Evaluation}
\label{sec:visual-ranking}

Twelve participants each ranked 20 sets of six anonymized four-minute videos according to their overall preference. AGF achieved the best mean rank of 2.23, followed by LongLive at 2.85 (Figure~\ref{fig:ranking-mean}). AGF also received the highest proportion of first-place rankings (35.8\%) and was ranked among the top three in 84.2\% of evaluations. Appendix~\ref{app:visual-ranking} provides a more detailed analysis together with the full rank distributions. Overall, these results show that AGF is consistently preferred by human evaluators for long-horizon video generation.

\subsection{Training Cost}
\label{sec:training-efficiency}

Although AGF adds perturbation selection and paired clean/adversarial training, its short schedule and five-second rollouts keep its training cost lower than most baselines.
Starting from the ODE initialization, AGF requires 400 outer iterations, including 80 generator updates, for a total of 364.4 GPU-hours on eight H20 GPUs.

For comparison, Self-Forcing requires 600 outer iterations and 431.6 GPU-hours. LongLive performs 3,000 iterations of long-video tuning with 60-second streaming sequences, requiring 2774.4 GPU-hours. OPSD-V uses approximately 45-second rollouts and supervises all four denoising steps after the warm-up blocks. Repeated student and teacher evaluations along these long rollouts raise its training cost to 944.4 GPU-hours.
 OPSD-V additionally requires real long-video data, whereas AGF post-training uses only text prompts and self-generated videos. Full training configurations and cost accounting are provided in Appendix~\ref{app:experiment-details}.

\FloatBarrier
\subsection{Ablation Studies}
\label{sec:ablations}

\setlength{\intextsep}{6pt}
\begin{table}[H]
  \centering
  \small
  \setlength{\abovecaptionskip}{0pt}
  \setlength{\belowcaptionskip}{4pt}
  \caption{\textbf{Ablation studies of AGF.}
  Panel (a) studies perturbation and regularization strategies, while panel (b) evaluates AGF under a different ODE initialization. Best scores in each row are bolded.}
  \label{tab:ablations}
  \setlength{\tabcolsep}{1pt}
  \renewcommand{\arraystretch}{1.05}
  \begin{tabular*}{\linewidth}{@{\extracolsep{\fill}}l*{8}{c}@{\hspace{12pt}}lcc@{}}
    \multicolumn{9}{c}{\textit{(a) Perturbation and regularization}} &
    \multicolumn{3}{c}{\textit{(b) ODE initialization}} \\[2pt]
    \cmidrule[\heavyrulewidth](r{6pt}){1-9}\cmidrule[\heavyrulewidth](l{6pt}){10-12}
    & Default & Perturbation & Regularizer &
    \multicolumn{3}{c}{Regularized steps} & \multicolumn{2}{c}{Reference model} &
    & \multicolumn{2}{c}{Model} \\
    \cmidrule(lr){2-2}\cmidrule(lr){3-3}\cmidrule(lr){4-4}
    \cmidrule(lr){5-7}\cmidrule(lr){8-9}\cmidrule(lr){11-12}
    Duration & \textbf{AGF} & Gaussian & None &
    1--2 & 1--3 & 1--4 & ODE init & Teacher &
    Duration & CF & $\mathrm{AGF}_{\mathrm{CF}}$ \\
    \cmidrule(r{6pt}){1-9}\cmidrule(l{6pt}){10-12}
    5\,s & 82.03 & 71.09 & 81.38 &
    77.75 & \textbf{82.06} & 74.30 & 78.79 & 80.78 &
    5\,s & \textbf{82.94} & 81.74 \\
    60\,s & \textbf{81.47} & 68.60 & 77.96 &
    76.24 & 80.40 & 72.77 & 75.63 & 79.74 &
    60\,s & 79.83 & \textbf{81.00} \\
    240\,s & \textbf{84.82} & 75.11 & 82.75 &
    79.87 & 83.94 & 78.19 & 79.72 & 82.82 &
    240\,s & 81.79 & \textbf{84.60} \\
    \cmidrule[\heavyrulewidth](r{6pt}){1-9}\cmidrule[\heavyrulewidth](l{6pt}){10-12}
  \end{tabular*}
\end{table}

\paragraph{Adversarial perturbations.}
We compare the selected history perturbations with Gaussian
perturbations of the same RMS magnitude. Adversarial selection improves
240-second Q5 by 9.71 pp, demonstrating the benefit of using gradients to select adversarial perturbations (Table~\ref{tab:ablations}(a)).

\paragraph{Regularization strategies.}
The prediction consistency regularizer is an integral component of AGF rather than an optional add-on. Removing it lowers 240-second Q5 from 84.82 to 82.75 and 60-second Total from 81.47 to 77.96. We attribute this drop to that without an anchor to the clean-history prediction, the adversarial branch can shift the global scene layout. Nevertheless, the unregularized variant still improves 240-second Q5 over Self-Forcing (82.47) and matches OPSD-V (82.84) with only 80 generator updates. This indicates that adversarial history selection alone already benefits long-horizon quality, while the regularizer is needed to preserve scene consistency.

We then study effects of the regularizer beyond the first denoising step. Regularizing the first two, three, or all four steps lowers 60-second Total and
240-second Q5, supporting the choice to guide the initial
prediction while leaving later steps free to refine details.

We then compare two alternative regularizers that use frozen reference models to guide the perturbed branch.
Specifically, the first denoising step is regularized toward either the prediction of the fixed ODE-initialized generator or an equivalent first-step velocity target derived from the frozen diffusion teacher. Both regularizers underperform the proposed regularizer at all three durations (Table~\ref{tab:ablations}(a)). Appendix~\ref{app:ablations} presents details on their formulation.

\paragraph{ODE initialization.}
To verify the effectiveness of AGF across different initialization strategies, we train
$\mathrm{AGF}_{\mathrm{CF}}$ from Causal Forcing's Stage 2 ODE
checkpoint. It outperforms the released Causal Forcing model by 2.81 pp
in 240-second Q5 (Table~\ref{tab:ablations}(b)).
Training details appear in Appendix~\ref{app:causal-init}.

\endgroup

\section{Conclusion}
\label{sec:conclusion}
Our work investigates long-horizon degradation in few-step autoregressive video generation and identifies the model's sensitivity to errors in its self-generated history as a key source of cumulative quality decay. Building on this observation, we introduce Adversarial Gradient Forcing (AGF), which constructs history perturbations that are particularly detrimental to subsequent generations and explicitly trains the model to remain robust to them. Despite being trained only on five-second rollouts, AGF sustains stable visual quality throughout four-minute generation.



\FloatBarrier

\bibliography{references}
\bibliographystyle{plainnat}

\clearpage
\beginappendix
\raggedbottom
\setlength{\floatsep}{8pt plus 2pt minus 2pt}
\setlength{\textfloatsep}{10pt plus 2pt minus 2pt}
\setlength{\intextsep}{8pt plus 2pt minus 2pt}
\setlength{\abovecaptionskip}{6pt}
\renewcommand{\floatpagefraction}{0.8}

\section{Local History Robustness and Adversarial Selection}
\label{app:local-history-theory}

This appendix explains how a history error affects later generated blocks
and derives the perturbation used by AGF. We begin by separating newly
introduced errors from errors carried forward through generation. This lets
us relate a controlled history perturbation to the same transitions. We then
explain why the size of a video difference alone is insufficient for choosing
an adverse perturbation, before deriving the direction selected by the DMD
gradient under the frame-sharing and RMS constraints.

\subsection{Causal Error Propagation}
\label{app:causal-error-propagation}
At each generation step, the model can both carry forward an existing error
and introduce a new one. To express these contributions separately, let
$S_k$ be a reference state after block $k$ and $\widetilde S_k$ the state
produced by the generator. Each state contains the history, cache, and other
variables needed to continue generation. With the sampler state and
randomness $\omega_{k+1}$ fixed, write the generator's transition as
$\widetilde S_{k+1}=\mathcal T_k(\widetilde S_k,\omega_{k+1})$. Define the
new error at the reference state by
\begin{equation}
  e_{k+1}
  =\mathcal T_k(S_k,\omega_{k+1})-S_{k+1}.
  \label{eq:new-rollout-error}
\end{equation}
Thus, $e_{k+1}$ is the error the model would introduce even if its input were
the reference state $S_k$. Its actual input may already differ from that
reference by $\Delta S_k=\widetilde S_k-S_k$, so we also need to account for
the effect of this existing difference.

Assume that the transition is continuously differentiable along the segment
joining the two states. Averaging its Jacobian along this segment gives
$A_k^{\mathrm{sec}}$, which maps the existing difference to its effect one
step later. Products of these operators describe propagation over several steps:
\begin{equation}
  A_k^{\mathrm{sec}}
  =\int_0^1D_S\mathcal T_k(S_k+\tau\Delta S_k,\omega_{k+1})\,d\tau,
  \qquad
  \Phi_{K,m}^{\mathrm{sec}}
  =A_{K-1}^{\mathrm{sec}}\cdots A_m^{\mathrm{sec}},
  \quad \Phi_{m,m}^{\mathrm{sec}}=I.
\end{equation}
Here $D_S\mathcal T_k$ is the Jacobian of the transition with respect to
its state input. The fundamental theorem of calculus gives the exact recursion
$\Delta S_{k+1}=A_k^{\mathrm{sec}}\Delta S_k+e_{k+1}$. Repeated substitution
then yields
\begin{equation}
  \Delta S_K
  =\Phi_{K,j}^{\mathrm{sec}}\Delta S_j
  +\sum_{m=j+1}^{K}\Phi_{K,m}^{\mathrm{sec}}e_m.
  \label{eq:causal-error-propagation}
\end{equation}
The first term is the effect of the history error already present at block
$j$. Each term in the sum is the effect of an error introduced at a later
step. In both cases, the transition product determines how that error
contributes to the final state difference.

We are interested in later video blocks, so we extract them from the full
state using $P_{>j,K}$, which selects blocks $j+1,\ldots,K$ at horizon $K$.
Define the gain from a state difference at step $m$ to these later blocks as
$\Lambda_{K,m}^{\mathrm{sec}}=\|P_{>j,K}\Phi_{K,m}^{\mathrm{sec}}\|_2$. Projecting the
exact recursion gives
\begin{equation}
  \|P_{>j,K}\Delta S_K\|_2
  \leq
  \Lambda_{K,j}^{\mathrm{sec}}\|\Delta S_j\|_2
  +\sum_{m=j+1}^{K}
  \Lambda_{K,m}^{\mathrm{sec}}\|e_m\|_2.
  \label{eq:finite-horizon-propagation-bound}
\end{equation}
Each contribution is therefore bounded by the size of the error multiplied
by its propagation gain. This separates the accuracy of an individual
generation step from the effect its error can have later.

Let $Y_j(\delta)$ denote the continuation obtained by autoregressive
generation after adding $\delta$ to history block $j$. We write
\begin{equation}
  \Delta Y_j(\delta)=Y_j(\delta)-Y_j(0)
  \label{eq:paired-history-response}
\end{equation}
for the difference between the perturbed and original continuations
with model parameters and sampling randomness fixed.
To express this comparison in terms of the state transitions, choose the
model's unperturbed rollout as the reference trajectory. Both branches then
follow the same transition with the same future randomness. Consequently,
$e_m=0$ for $m>j$, and the perturbation supplies the initial difference
$\Delta S_j=S_j(\delta)-S_j(0)$. Projecting the resulting state difference
onto future outputs gives
\[
  \Delta Y_j(\delta)
  =P_{>j,K}\Phi_{K,j}^{\mathrm{sec}}
    \bigl[S_j(\delta)-S_j(0)\bigr].
\]
Here every later difference comes from the single history intervention.
The reference is the model's own rollout, so setting these comparison
residuals to zero does not assume that the reference video is free of artifacts.

\subsection{Why Video Differences Need a Quality Criterion}
\label{app:distribution-response}
The preceding analysis describes how much later outputs change. To select
training perturbations, however, we also need to assess whether those changes
are adverse to distribution matching. The following example shows why these
are different questions: a generator can match its target distribution
exactly while its outputs remain highly sensitive to history under shared noise.

Consider a target conditional distribution $\mathcal N(0,I_2)$ at every
history $h\in\mathbb R$. Let $z\sim\mathcal N(0,I_2)$, let
$\operatorname{Rot}(\alpha)$ be a two-dimensional rotation, and define the
smooth family of generators
\begin{equation}
  Q_\kappa(h,z)=\operatorname{Rot}(\kappa h)z.
  \label{eq:distribution-response-separation}
\end{equation}
Rotational invariance gives
$\operatorname{Law}(Q_\kappa(h,z)\mid h)=\mathcal N(0,I_2)$ for every $h$ and
$\kappa$, so the conditional distribution is matched independently of
$\kappa$. With the same $z$ at two neighboring histories,
\begin{equation}
  \mathbb E\left\|\partial_hQ_\kappa(h,z)\right\|_2^2=2\kappa^2,
  \qquad
  \mathbb E\left\|Q_\kappa(h+\delta,z)-Q_\kappa(h,z)\right\|_2^2
  =4\bigl(1-\cos(\kappa\delta)\bigr).
  \label{eq:unbounded-paired-sensitivity}
\end{equation}

Increasing $\kappa$ makes the local derivative arbitrarily large without
changing the conditional output distribution. Thus, large sensitivity under
shared randomness alone does not establish a distribution-matching error.
With exact score estimates, the target and generated scores in this example
coincide, so the DMD gradient is zero for every $\kappa$. This is why AGF uses
the DMD gradient to assess a history perturbation's effect on later blocks,
rather than selecting perturbations solely by the size of the resulting
video difference.

\subsection{Backpropagating the Future DMD Gradient}
\label{app:tangent-pullback}
AGF selects the perturbation using the parallel forward pass described in
Section~\ref{sec:history-direction}(b). This pass uses the recorded noisy
inputs $\bar X_{t^\star}$ at $t^\star$ and the recorded clean latent blocks $\bar X_0$
as history, with $\delta$ added only to $\bar x_0^{(j)}$.
Let $\widehat y_j^{(k)}(\delta)$ denote the clean latent prediction for
block $k$, and collect the predictions for blocks after $j$ as
$\widehat Y_j(\delta)=(\widehat y_j^{(j+1)}(\delta),\ldots,\widehat y_j^{(K)}(\delta))$.
The recorded inputs, all other history blocks, and model parameters remain
fixed. Let $J_{j,k}$ and $J_j$ denote the Jacobians of
$\widehat y_j^{(k)}(\delta)$ and $\widehat Y_j(\delta)$ with respect to
$\delta$, evaluated at $\delta=0$.
Writing $\bar g_{j,k}$ for the entries of the fixed DMD gradient $\bar g_j$
corresponding to block $k$, backpropagation gives
\begin{equation}
  u_j
  =\sum_{k=j+1}^{K}J_{j,k}^{\top}\bar g_{j,k}
  =J_j^\top\bar g_j.
  \label{eq:causal-pullback-products}
\end{equation}
Each term maps the DMD gradient of a later block to a gradient on history
block $j$ through the differentiable history K/V representations. A single
vector--Jacobian product computes their sum without forming the Jacobians.
These derivatives describe predictions computed from recorded histories.
Appendix~\ref{app:causal-error-propagation} describes error propagation
during autoregressive generation, where newly generated blocks become
history for subsequent blocks.

\subsection{Selecting a Perturbation under Sharing and RMS Constraints}
\label{app:pullback}

Once we have a gradient with respect to the history block, we need to choose
a perturbation that satisfies the frame-sharing and RMS constraints.
Using the Jacobian $J_j$ defined above, let
$\Delta\widehat Y_j(\delta)=\widehat Y_j(\delta)-\widehat Y_j(0)$.
The criterion $\phi_j(\delta)=\inner{\bar g_j}{\Delta\widehat Y_j(\delta)}$
measures the prediction change along the fixed DMD gradient.
Holding $\bar g_j$ fixed gives
\begin{equation}
  \phi_j(\delta)
  =\inner{\bar g_j}{J_j\delta}+o(\|\delta\|)
  =\inner{J_j^\top\bar g_j}{\delta}+o(\|\delta\|).
  \label{eq:pullback-expansion}
\end{equation}
The linear term is therefore maximized by aligning the perturbation with
$u_j=J_j^\top\bar g_j$, subject to those constraints. For $T_j$ latent frames
and flattened spatial-channel dimension $D$, the allowed set is
\begin{equation}
  \begin{aligned}
    \mathcal S_{\mathrm{share},j}
    &=\left\{\delta\in\mathbb R^{T_j\times D}:
      \delta_1=\cdots=\delta_{T_j}\right\},\\
    \RMS(\delta)
    &=\sqrt{\frac{1}{T_jD}
      \sum_{\tau=1}^{T_j}\sum_{d=1}^{D}\delta_{\tau,d}^2},\\
    \mathcal B_{\epsilon,j}
    &=\left\{\delta\in\mathcal S_{\mathrm{share},j}:
      \RMS(\delta)\leq\epsilon\right\}.
  \end{aligned}
  \label{eq:rms-threat-model}
\end{equation}
Here $\mathcal S_{\mathrm{share},j}$ requires the same perturbation in every
frame, and $\mathcal B_{\epsilon,j}$ also bounds its RMS magnitude. To enforce
sharing, the orthogonal projection averages the history gradient over frames
and assigns the same average to every frame. This gives the shared
direction $\widetilde u_j=\Pi_{\mathrm{share}}(u_j)$ used in the main text:
\begin{equation}
  [\widetilde u_j]_\tau
  =[\Pi_{\mathrm{share}}(u_j)]_\tau
  =\frac{1}{T_j}\sum_{\ell=1}^{T_j}[u_j]_\ell,
  \qquad \tau=1,\ldots,T_j.
  \label{eq:time-shared-projection}
\end{equation}
For every $\delta\in\mathcal S_{\mathrm{share},j}$,
$\inner{u_j}{\delta}=\inner{\widetilde u_j}{\delta}$. Cauchy--Schwarz
therefore gives
\begin{equation}
  \argmax_{\delta\in\mathcal B_{\epsilon,j}}
    \inner{u_j}{\delta}
  =\epsilon\,
    \frac{\widetilde u_j}{\RMS(\widetilde u_j)}
  \quad\text{when }\widetilde u_j\neq0.
  \label{eq:time-shared-support-point}
\end{equation}
Stopping gradients through this normalized direction gives $\delta_j^\star$
in Section~\ref{sec:history-direction}(b). When $\widetilde u_j=0$,
every allowed perturbation has the same linearized value, and AGF
chooses $\delta_j^\star=0$. The optimality here applies to the first-order
criterion with $\bar g_j$ fixed; the finite perturbation is subsequently
used to recompute predictions under the perturbed history for generator training.

\section{Paired Training and Gradient Computation}
\label{app:paired-training}

AGF first differentiates with respect to a history perturbation, then fixes
that perturbation and differentiates with respect to the generator parameters.
This appendix specifies the DMD gradients and the computation paths for these
two operations. It also gives the fake-score update, regularizer weighting, and
complete training algorithm.

\subsection{Applying DMD Gradients to the Generator}
\label{app:gradient-carrier}

Let $X_0=(x_0^{(1)},\ldots,x_0^{(K)})$ denote a full video represented by its
clean latent blocks, and let $c$ be its text prompt.
We use $q(X_0\mid c)$ for the distribution
of videos generated from the original and perturbed histories, weighted
according to the two branch losses, and
$p_{\mathrm{real}}(X_0\mid c)$ for the target video distribution.
DMD compares these distributions after adding Gaussian noise to the videos
\citep{yin2024dmd,yin2024dmd2}.
At diffusion time $\tau$, the noisy video is
$X_\tau=\alpha_\tau X_0+\sigma_\tau\varepsilon_{\mathrm{diff}}$, where
$\varepsilon_{\mathrm{diff}}\sim\mathcal N(0,I)$ and
$\alpha_\tau,\sigma_\tau$ specify the noise schedule. Writing the corresponding
noisy distributions as $q_\tau$ and $p_{\mathrm{real},\tau}$, the DMD objective is
\begin{equation}
  \mathcal L_{\mathrm{DM}}(q)
  =\mathbb E_{c,\tau}\!\left[
    D_{\mathrm{KL}}\!\left(
      q_\tau(\cdot\mid c)\,\|\,p_{\mathrm{real},\tau}(\cdot\mid c)
    \right)
  \right].
  \label{eq:dm-distribution-objective}
\end{equation}

To make the sampling explicit, we write the DMD gradient introduced in
Section~\ref{sec:history-direction} as $g_{\mathrm{DM}}(X_0;\xi)$ in this appendix,
where $\xi=(\tau,\varepsilon_{\mathrm{diff}})$ contains the sampled diffusion
time and noise. These are sampled separately from the randomness used to
generate the video, including the denoising step $t^\star$ selected for
the generator's recorded inputs $\bar X_{t^\star}$.
Overbars denote values held fixed during differentiation.
Equation~\ref{eq:dm-distribution-objective} defines the distributional
objective. To derive the gradient used to train the generator,
consider a differentiable sampler
$X_{0,\theta}\sim q_\theta(\cdot\mid c)$ and define the scores of the noisy
generated and target distributions by
\[
  s_{q_\theta}(X_\tau,\tau,c)=\nabla_{X_\tau}\log q_{\theta,\tau}(X_\tau\mid c),
  \qquad
  s_{\mathrm{real}}(X_\tau,\tau,c)
  =\nabla_{X_\tau}\log p_{\mathrm{real},\tau}(X_\tau\mid c).
\]
Using exact scores, the distribution-matching gradient is
\citep{yin2024dmd,yin2024dmd2}
\begin{equation}
  \nabla_\theta\mathcal L_{\mathrm{DM}}(q_\theta)
  =\mathbb E\!\left[
    (D_\theta X_{0,\theta})^\top\alpha_\tau
    \bigl(s_{q_\theta}(X_\tau,\tau,c)-s_{\mathrm{real}}(X_\tau,\tau,c)\bigr)
  \right].
  \label{eq:dm-score-gradient}
\end{equation}
Here $D_\theta X_{0,\theta}$ is the Jacobian of the generated video with respect
to the generator parameters. The expectation includes prompts, generator
sampling randomness, and the diffusion time and noise. The factor $\alpha_\tau$ comes from differentiating
$X_\tau=\alpha_\tau X_{0,\theta}+\sigma_\tau\varepsilon_{\mathrm{diff}}$ with respect
to the generated video. Thus, the difference between the two scores supplies
an output gradient that is backpropagated through the generator.

For the Gaussian noising process, a denoised video prediction and its score
estimate satisfy
\[
  \widehat s_a(X_\tau,\tau,c)
  =\frac{\alpha_\tau\widehat X_{0,a}(X_\tau,\tau,c)-X_\tau}{\sigma_\tau^2},
  \qquad a\in\{\mathrm{real},\mathrm{fake}\}.
\]
Substituting these estimates into the score difference gives
$
  \alpha_\tau(\widehat s_{\mathrm{fake}}-\widehat s_{\mathrm{real}})
  =\frac{\alpha_\tau^2}{\sigma_\tau^2}
    (\widehat X_{0,\mathrm{fake}}-\widehat X_{0,\mathrm{real}})$.
The practical DMD gradient uses estimated scores and an additional
normalization:
\begin{equation}
  g_{\mathrm{DM}}(X_0;\xi)
  =\gamma_{\mathrm{DM}}(X_0,\tau)\left[
    \widehat X_{0,\mathrm{fake}}(X_\tau,\tau,c)
    -\widehat X_{0,\mathrm{real}}(X_\tau,\tau,c)
  \right],
  \label{eq:dmd-field}
\end{equation}
where $\gamma_{\mathrm{DM}}(X_0,\tau)>0$ includes diffusion scaling and
normalization. When this
normalization depends on the sample, it changes the weighting of samples
relative to the unscaled KL gradient in Eq.~\ref{eq:dm-score-gradient}.

To backpropagate a fixed DMD gradient $g$ through the generator without
differentiating through either score model, we construct the mean squared error
surrogate loss
\begin{equation}
  \ell_{\mathrm{DM}}(X_0;g)
  =\frac{1}{2N}\left\|X_0-\stopgrad(X_0-g)\right\|_2^2,
  \qquad \nabla_{X_0}\ell_{\mathrm{DM}}(X_0;g)=\frac{g}{N}.
  \label{eq:dmd-carrier}
\end{equation}
Here $N$ is the number of scalar latent coordinates in $X_0$. The target
$\stopgrad(X_0-g)$ is treated as constant, so gradients pass only through the
first occurrence of $X_0$. Differentiation therefore gives
$(X_0-\stopgrad(X_0-g))/N=g/N$, with $1/N$ implementing the mean reduction.
The loss thus applies the fixed DMD gradient, with a factor of $1/N$ from
averaging over latent coordinates; its numerical value is not
an estimate of the KL divergence.

During perturbation selection, we use the DMD gradient before this averaging
and keep only its entries for blocks after $j$:
\begin{equation}
  \bar g_j
  =\stopgrad\!\left(
    [g_{\mathrm{DM}}(X_{0,j}^{\mathrm{in}}(0);\xi^{\mathrm{clean}})]_{>j}
  \right),
  \label{eq:stopped-future-field}
\end{equation}
Here $X_{0,j}^{\mathrm{in}}(0)$ is the full video prediction recomputed under
the original history from the recorded inputs, as defined in
Appendix~\ref{app:gradient-boundaries},
and $[\cdot]_{>j}$ extracts its later blocks. During generator training,
each branch $b\in\{\mathrm{clean},\mathrm{adv}\}$ uses
$g^b=\stopgrad(g_{\mathrm{DM}}(X_0^b;\xi^b))$ evaluated on its own full
prediction $X_0^b$. The clean branch reuses the DMD gradient computed for
selection; the adversarial branch draws its own diffusion time and noise.
These DMD draws are separate from the recorded noisy inputs used for
the parallel forward pass.

The adversarial loss averages the same squared error only over blocks after
$j$. Let $N_{>j}$ be their number of scalar latent coordinates. Then
\[
  \ell_{\mathrm{DM}}^{>j}(X_0;g)
  =\frac{1}{2N_{>j}}\left\|
    [X_0]_{>j}-\stopgrad\!\left([X_0]_{>j}-[g]_{>j}\right)
  \right\|_2^2.
\]
Its gradient with respect to $X_0$ is zero on blocks up to $j$ and equals
$[g]_{>j}/N_{>j}$ on later blocks. The two branch losses therefore average
over different coordinate sets; their weights in Eq.~\ref{eq:agf-objective}
apply to these mean losses.
For perturbation selection, multiplying the mean-reduced inner loss by
$N_{>j}$ recovers the unaveraged gradient $J_j^\top\bar g_j$.
At fixed $j$, this positive scaling does not change the direction after RMS
normalization. The mask determines where gradients are applied; DMD itself
is evaluated on the full video in both selection and training.

\subsection{Parallel Forward Passes for Selection and Training}
\label{app:gradient-boundaries}

The autoregressive rollout in Section~\ref{sec:history-direction}(a)
records the noisy inputs $\bar X_{t^\star}$ at a sampled denoising step
$t^\star$ shared across all blocks and their corresponding clean predictions
$\bar X_0$, without retaining the computation graph. These inputs are recorded
from the generator's denoising trajectory; they are not constructed by
adding fresh Gaussian noise to $\bar X_0$. The independent forward-noising
relation in Appendix~\ref{app:gradient-carrier} instead applies to a clean
video prediction $X_0$ and its DMD input $X_\tau$.
The text prompt is denoted by $c$.
Let $\bar X_0(\delta)$ be $\bar X_0$ with $\delta$ added only to $\bar x_0^{(j)}$.
We write $\mathcal R_\theta(\bar X_{t^\star},t^\star;\bar X_0(\delta),c)$ for the full
video prediction produced by the generator's parallel forward pass.

The generator encodes the recorded clean latent blocks and noisy inputs
together under causal attention. Each noisy block uses the K/V
representations of the preceding recorded blocks as history.
Since these history blocks are already available, all block predictions
can be computed in parallel. Newly computed predictions are not inserted
into the history during this pass. The serial K/V cache is released after
the rollout; each parallel forward pass recomputes differentiable history
K/V representations from the recorded clean latent blocks.

During selection, the parameters $\bar\theta=\stopgrad(\theta)$ are fixed,
and the zero-initialized $\delta$ is the variable being differentiated:
\begin{equation}
  X_{0,j}^{\mathrm{in}}(\delta)
  =\mathcal R_{\bar\theta}(\bar X_{t^\star},t^\star;\bar X_0(\delta),c),
  \qquad
  \widehat Y_j(\delta)=[X_{0,j}^{\mathrm{in}}(\delta)]_{>j}.
  \label{eq:selection-prediction}
\end{equation}
DMD is evaluated on the complete prediction $X_{0,j}^{\mathrm{in}}(0)$.
Its later-block entries are backpropagated through the parallel forward pass to
$\delta$, producing the gradient used to construct $\delta_j^\star$.

We then detach the selected perturbation and discard the selection graph.
For $b\in\{\mathrm{clean},\mathrm{adv}\}$, let
$\delta^{\mathrm{clean}}=0$ and $\delta^{\mathrm{adv}}=\delta_j^\star$.
The fixed clean latent blocks used as history in each branch are
$C^b=\stopgrad(\bar X_0(\delta^b))$. Using the same parallel computation,
the clean and adversarial branches produce
\begin{align}
  X_0^b
  &=\mathcal R_\theta\!\left(
    \bar X_{t^\star},t^\star;C^b,c\right),\nonumber\\
  g^b&=\stopgrad(g_{\mathrm{DM}}(X_0^b;\xi^b)).
  \label{eq:paired-predictions}
\end{align}
The recorded noisy inputs and history blocks are fixed, but the history K/V
representations are recomputed with trainable parameters. Gradients therefore
update both history encoding and denoising. Each $X_0^b$ contains predictions
for all video blocks. Before the parameter update, the clean branch uses
the same recorded inputs, history, and parameters as perturbation selection, so
\begin{equation}
  X_0^{\mathrm{clean}}=X_{0,j}^{\mathrm{in}}(0).
  \label{eq:zero-consistency}
\end{equation}
This permits the clean branch to reuse the DMD gradient computed during selection.
Table~\ref{tab:gradient-boundaries} summarizes the two gradient computations.

\begin{table}[ht]
  \centering
  \small
  \caption{Gradient paths for perturbation selection and generator training.}
  \label{tab:gradient-boundaries}
  \begin{tabular}{@{}>{\raggedright\arraybackslash}p{0.21\linewidth}>{\raggedright\arraybackslash}p{0.33\linewidth}>{\raggedright\arraybackslash}p{0.40\linewidth}@{}}
    \toprule
    & Perturbation selection & Generator training \\
    \midrule
    Differentiated variable & history perturbation $\delta$ & generator parameters $\theta$ \\
    Active path & history K/V computation and later predictions & history K/V computation and predictions \\
    Fixed values & parameters, recorded inputs and history, DMD gradient & perturbation, recorded inputs and history, DMD gradient \\
    Output & history perturbation & generator update \\
    \bottomrule
  \end{tabular}
\end{table}

\FloatBarrier
\subsection{Training the Fake-Score Model}
\label{app:fake-score}

The generator learns from full video predictions recomputed under the
original and perturbed histories. We train one fake-score model on their
weighted mixture. With branch loss weights
$\lambda_{\mathrm{clean}}$ and $\lambda_{\mathrm{adv}}$, define
\begin{equation}
  \pi_{\mathrm{clean}}
  =\frac{\lambda_{\mathrm{clean}}}
          {\lambda_{\mathrm{clean}}+\lambda_{\mathrm{adv}}},
  \qquad
  q_{\mathrm{mix}}^{(s)}
  =\pi_{\mathrm{clean}}q_{\mathrm{clean}}^{(s)}
   +(1-\pi_{\mathrm{clean}})q_{\mathrm{adv}}^{(s)}.
  \label{eq:branch-mixture}
\end{equation}
Here $q_{\mathrm{clean}}^{(s)}$ and $q_{\mathrm{adv}}^{(s)}$ are the video
distributions from the two branches at training iteration $s$. They use the
same sampling law over prompts, generator noise, history positions, and DMD
draws $(c,\omega,j,\xi)$. Both branch weights are $0.5$ in the default configuration.
The fake-score model is shared by the clean and adversarial branches.
For each branch, the same teacher and fake-score model are evaluated on
the Gaussian-noised version of that branch's full video prediction.

Training also reuses generated videos stored in a replay buffer. If
$N_{\mathrm{replay}}$ is the maximum age and $\beta_\ell$ is the sampling
weight for videos of age $\ell$, their distribution is
\begin{equation}
  q_{\mathrm{replay}}^{(s)}
  =\sum_{\ell=0}^{N_{\mathrm{replay}}}
    \beta_\ell q_{\mathrm{mix}}^{(s-\ell)},
  \qquad \beta_\ell\geq0,\quad
  \sum_{\ell=0}^{N_{\mathrm{replay}}}\beta_\ell=1.
  \label{eq:replay-mixture}
\end{equation}
The default replay scheme preserves the branch proportions at every age. All videos
are detached from the generator before the fake-score model is trained on
them with the diffusion regression objective used by DMD. Thus, this
update trains the score estimate without updating the video generator.

\subsection{Prediction Consistency Regularization and Weighting}
\label{app:prediction-consistency-details}

The regularizer compares first-step velocity predictions under the two histories
used in Eq.~\ref{eq:paired-predictions}. Let $V_\theta$ denote parallel velocity
prediction at the highest-noise training timestep and $Z_{\max}$ the
initial noisy latents recorded during sampling. Writing $C^b$ for the fixed
clean latent blocks used as history in branch $b$, the predictions for
blocks after $j$ are
\[
  v_{j,1}^b
  =[V_\theta(Z_{\max};C^b,c)]_{>j},
  \qquad b\in\{\mathrm{clean},\mathrm{adv}\}.
\]
Both calls use the current generator parameters and the same noisy inputs.
We recompute the clean prediction for each update and then detach it to
provide the reference in Eq.~\ref{eq:prediction-consistency}.
The operator $\operatorname{MSE}_{>j}$ averages squared differences over
coordinates after block $j$. Gradients pass through the adversarial
prediction and its history K/V computation; the recorded history blocks and
clean reference remain fixed.

The gradients from DMD and the regularizer can change scale during training. To control
their relative strength, we weight the regularizer using the ratio of estimated
gradient RMS values at the model outputs. At iteration $s$, denote these estimates by
$r_{\mathrm{DM}}^{(s)}$ and $r_{\mathrm{reg}}^{(s)}$, and let $\rho_s$ be
the desired relative strength of the regularizer. The loss weight is
\begin{equation}
  \lambda_s
  =\stopgrad\!\left(
    \min\!\left\{\lambda_{\max},
    \rho_s\frac{r_{\mathrm{DM}}^{(s)}}
    {r_{\mathrm{reg}}^{(s)}+\epsilon_{\mathrm{num}}}\right\}
  \right),
  \qquad
  \mathcal L_{\mathrm{train}}
  =\mathcal L_{\mathrm{AGF}}+\lambda_s\mathcal L_{\mathrm{reg}}.
  \label{eq:regularizer-scaling}
\end{equation}
The ratio adjusts the regularization coefficient as the two gradient scales change;
the cap $\lambda_{\max}$ and stabilizer $\epsilon_{\mathrm{num}}$ limit
large weights when the regularizer gradient estimate is small. These are estimates at
the outputs used for calibration, rather than norms of the full generator
parameter gradients. The coefficient itself is detached during the update.

The evaluated configuration uses $\epsilon_{\mathrm{num}}=10^{-12}$ and
$\lambda_{\max}=1000$. The target ratio $\rho_s$ rises from zero at
iteration 20 to $0.05$ at iteration 80, stays there through iteration 200,
and is zero for iterations 201--400. Thus, $0.05$ is the target gradient
ratio, not a fixed loss coefficient. After the regularizer is removed, training
continues with Eq.~\ref{eq:agf-objective}. Table~\ref{tab:training-config}
records the complete evaluated configuration.

\subsection{Complete Training Update}
\label{app:agf-algorithm}

Algorithm~\ref{alg:agf} collects the computations for a generator update.
The fake-score model is trained separately using videos from both branches,
detached from the generator and stored in replay.
In our experiments, there are five fake-score updates per generator
update; the 400 outer iterations therefore contain 80 generator updates
and 400 fake-score updates. The regularization schedule uses the outer iteration
index $s$.

\begin{algorithm}[ht]
  \caption{AGF generator update}
  \label{alg:agf}
  \begin{algorithmic}[1]
    \REQUIRE Generator $\theta$, prompt $c$, perturbation budget $\epsilon$,
    learning rate $\eta_\theta$, branch weights
    $\lambda_{\mathrm{clean}},\lambda_{\mathrm{adv}}$, iteration $s$.
    \STATE Perform an autoregressive rollout without retaining its graph.
    Record noisy inputs $\bar X_{t^\star}$ at a sampled denoising step
    $t^\star$ shared across all blocks and corresponding clean predictions
    $\bar X_0$; retain the
    initial noise $Z_{\max}$ for regularization. Select $j<K$.
    \STATE With $\theta$ fixed, recompute the full video prediction
    $X_{0,j}^{\mathrm{in}}(\delta)$ at $\delta=0$ with a parallel forward pass
    (Eq.~\ref{eq:selection-prediction}).
    \STATE Sample $\xi^{\mathrm{clean}}$ and evaluate the full-video DMD gradient
    $g^{\mathrm{clean}}\leftarrow\stopgrad(g_{\mathrm{DM}}(X_{0,j}^{\mathrm{in}}(0);\xi^{\mathrm{clean}}))$.
    Set $\bar g_j\leftarrow[g^{\mathrm{clean}}]_{>j}$.
    \STATE Backpropagate to history:
    $u_j\leftarrow J_j^\top\bar g_j$
    (Eq.~\ref{eq:causal-pullback-products}); form the shared direction
    $\widetilde u_j\leftarrow\Pi_{\mathrm{share}}(u_j)$
    (Eq.~\ref{eq:time-shared-projection}).
    \STATE Set $\delta_j^\star\leftarrow
    \stopgrad(\epsilon\widetilde u_j/\RMS(\widetilde u_j))$, or zero if
    $\widetilde u_j=0$. Discard the selection graph.
    \STATE Add $\delta_j^\star$ to $\bar x_0^{(j)}$ to form the perturbed history.
    Recompute $X_0^{\mathrm{clean}}$ and $X_0^{\mathrm{adv}}$ under the original
    and perturbed histories with $\theta$ trainable and recorded inputs
    and history blocks fixed (Eq.~\ref{eq:paired-predictions}).
    \STATE Reuse $g^{\mathrm{clean}}$, sample $\xi^{\mathrm{adv}}$, and evaluate
    $g^{\mathrm{adv}}\leftarrow\stopgrad(g_{\mathrm{DM}}(X_0^{\mathrm{adv}};\xi^{\mathrm{adv}}))$.
    Form $\mathcal L_{\mathrm{AGF}}$ (Eq.~\ref{eq:agf-objective}).
    \STATE During the regularization phase, compute the paired first-step velocity
    predictions and $\mathcal L_{\mathrm{reg}}$
    (Eq.~\ref{eq:prediction-consistency}); set $\lambda_s$ using
    Eq.~\ref{eq:regularizer-scaling}. Otherwise set $\lambda_s=0$.
    \STATE Update the generator through history K/V computation and denoising:
    $\theta\leftarrow\theta-\eta_\theta\nabla_\theta
    (\mathcal L_{\mathrm{AGF}}+\lambda_s\mathcal L_{\mathrm{reg}})$.
    \STATE Store detached branch videos in replay for fake-score training
    with proportions $\lambda_{\mathrm{clean}}:\lambda_{\mathrm{adv}}$
    (Appendix~\ref{app:fake-score}).
  \end{algorithmic}
\end{algorithm}

\FloatBarrier
\section{Visualizing Changes in History and Continuations}
\label{app:gradient-phenotype-gallery}

\begin{figure}[!htbp]
  \centering
  \includegraphics[width=0.94\linewidth]{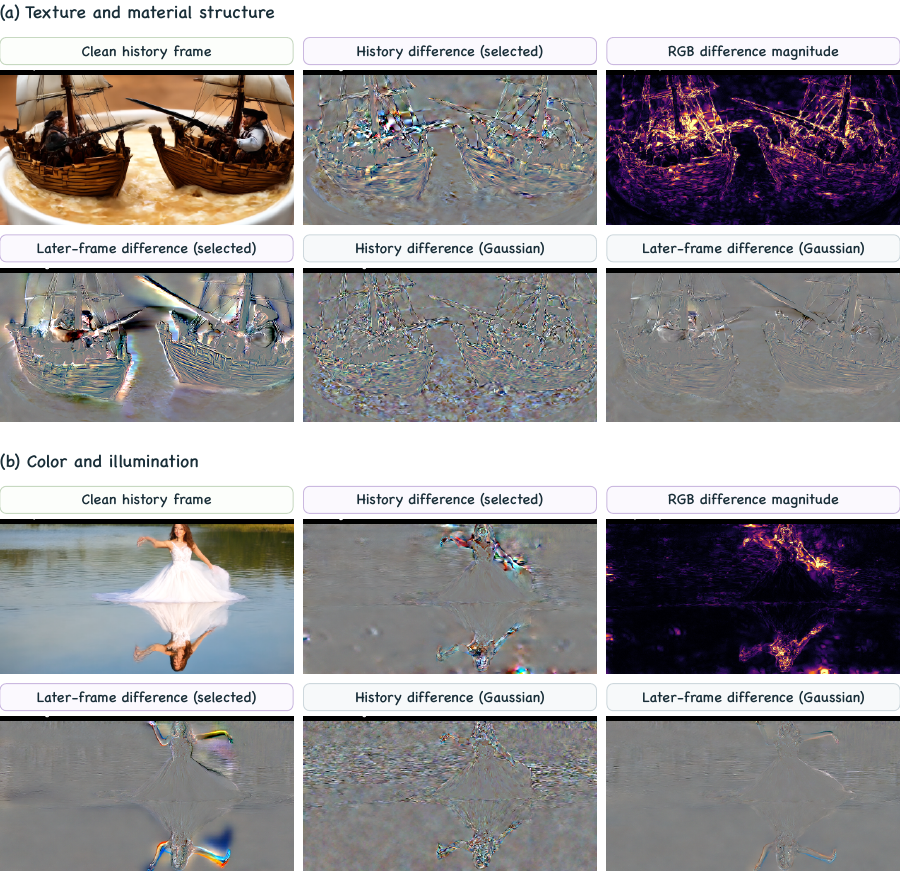}
  \caption{\textbf{Pixel changes around objects and illumination regions.}
  In the ship scene, the selected perturbation changes pixels along the
  hulls, masts, and material boundaries. In the dancer scene, changes appear
  around the subject, her reflection, and regions of similar brightness.
  Similar patterns appear in the differences between later frames. The
  matched Gaussian perturbation produces weaker and more diffuse
  later-frame differences in these examples.}
  \label{fig:gradient-phenotype-gallery-appearance}
\end{figure}

\begin{figure}[!htb]
  \centering
  \includegraphics[width=0.94\linewidth]{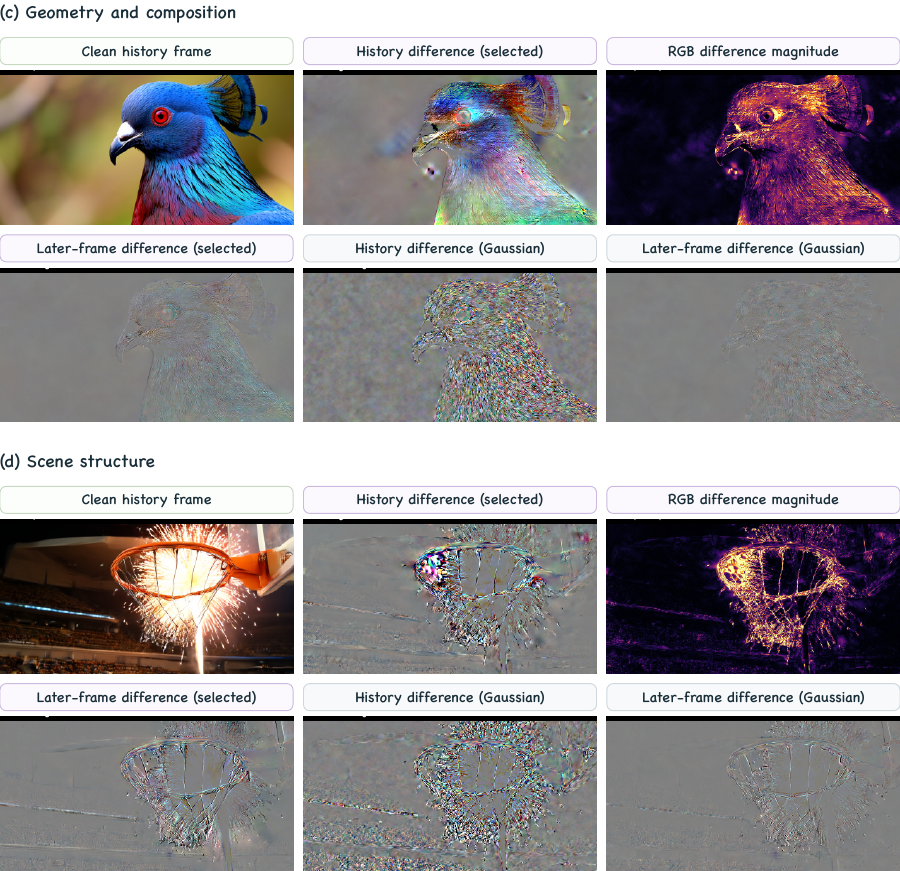}
  \caption{\textbf{Pixel changes along object outlines and scene structures.}
  The selected perturbations change pixels along the pigeon's head and body
  outline, or around the hoop, net, and surrounding basketball scene.
  Similar contours appear in the differences between later frames, while
  the matched Gaussian perturbation produces more diffuse changes.}
  \label{fig:gradient-phenotype-gallery}
\end{figure}

The figures below show where a history perturbation changes the decoded
video. For each example, we decode the clean and perturbed videos separately
and subtract their RGB pixels. We show this difference in the history block
and in a later generated frame, so the reader can compare the initial change
with its effect on subsequent generation. The clean frame provides a scene
reference.

Figure~\ref{fig:gradient-phenotype} uses an archived diagnostic checkpoint to inspect the perturbation-selection procedure. Its panels
show the same scene and perturbations as the ship example below. For each
comparison, the model parameters, original history, and future sampling
randomness are fixed; only the inserted perturbation changes.

History panels use frame 27, within the perturbed block, and later-frame
panels use frame 48; indices are zero-based. We compare the selected
perturbation with Gaussian noise at the same latent RMS budget, with each
pattern shared across three latent frames.

Let $\Delta I=I_{\mathrm{perturbed}}-I_{\mathrm{clean}}$, where both decoded
RGB images have values in $[0,1]$. To make small differences visible, all
signed maps use the display transform
$\operatorname{clip}_{[0,1]}(0.5+4\Delta I)$; mid-gray denotes zero difference. Magnitude
maps average the absolute RGB differences over channels, amplify them
$8\times$, and clip them before applying the Inferno colormap. All panels
use these same scales. The RMS budget is matched in latent space before
decoding; the maps show differences between decoded videos.

\section{Training and Evaluation Details}
\label{app:experiment-details}

\subsection{Training Configuration}

Table~\ref{tab:training-config} summarizes the configuration of the model
evaluated in Section~\ref{sec:experiments}, using its EMA weights at iteration
400. Training resumes at iteration 200 with the regularizer disabled, retaining
the generator, fake-score model, optimizer, EMA, replay buffer, random-number
generator, and data-loader states.

\begin{table}[!htbp]
  \centering
  \small
  \caption{Training configuration for the evaluated AGF model. Optimizer and
  perturbation settings correspond to the pure-AGF stage.}
  \label{tab:training-config}
  \begin{tabular}{p{0.29\linewidth}p{0.65\linewidth}}
    \toprule
    Field & Configuration \\
    \midrule
    Generator / fake-score model & Wan2.1-T2V-1.3B / one Wan2.1-T2V-1.3B fake-score model \\
    Frozen target score model & Wan2.1-T2V-14B \\
    Initialization & Self-Forcing ODE checkpoint \\
    Prompt source & Self-Forcing's filtered, LLM-extended VidProM text prompts; sampled line-wise \\
    Training rollout & 21 latent frames; 3 frames per block; 7 latent blocks ($\sim$5 s) \\
    Steps 1--200 & AGF with regularization applied to the first denoising prediction \\
    Regularization schedule & $\rho_s=0$ at step 20, linear ramp to $0.05$ at step 80, hold through step 200 \\
    Resuming training & Full training state resumed at step 200; optimizer and replay retained \\
    Steps 201--400 & Pure AGF; the regularizer disabled; step-400 EMA evaluated \\
    Perturbation & One gradient step using DMD gradients on later blocks; candidate block indices $[1,2,3]$; RMS budget 0.05; same pattern in every latent frame; no random initialization \\
    Branch weights & Clean/adversarial generator loss $0.5/0.5$ \\
    Fake-score updates & One fake-score model; $5{:}1$ ratio of fake-score to generator updates \\
    Replay & Enabled; 8 batches per rank on CPU in bf16; clean/adversarial $0.5/0.5$ \\
    Batch & Microbatch 1 per rank; 8 ranks; accumulation 8; effective global batch 64 \\
    Optimizer & AdamW, $\beta=(0,0.999)$; generator/fake-score LR $2\!\times\!10^{-6}/4\!\times\!10^{-7}$; generator weight decay 0.01 \\
    Sampler in training & Nodes $[1000,750,500,250]$ mapped through the scheduler with shift 5.0; same sampled exit step across blocks; guidance 3.0 \\
    Gradient / precision & Max generator/fake-score norm 10/10; bf16; gradient checkpointing \\
    Distributed training & FSDP \texttt{hybrid\_full}; 8$\times$ NVIDIA H20; seed 20260731 \\
    EMA & Weight 0.99; EMA restored at step 200 and evaluated at step 400 \\
    \bottomrule
  \end{tabular}
\end{table}

\paragraph{Training-cost accounting.}
Table~\ref{tab:training-cost} lists the training configuration, video-data
requirements, and timing measurements for the stage included in each method's
cost estimate. All timings use eight
H20 GPUs. Iterations count outer training loops: AGF's 400 iterations comprise
80 generator updates and 400 fake-score updates.

For methods whose costs are extrapolated, the budget is
$8N\bar t(B/B_{\mathrm{timed}})/3600$ GPU-hours, where $N$ and $B$ are the
iterations and global batch in Table~\ref{tab:training-cost}(a), and $\bar t$
is the mean seconds per iteration at batch $B_{\mathrm{timed}}$ in (b).
This assumes linear workload scaling with batch. Self-Forcing uses historical
rank-0 log timestamps; OPSD-V and LongLive use synchronized timings maximized
over all eight ranks. For DMD-based methods, timings are averaged over
complete cycles of generator and fake-score updates. OPSD-V timing uses
synthetic inputs with 180 latent frames;
its target batch of 24 follows the reported 24 devices with one sample each
and assumes no extra accumulation.

AGF's recorded elapsed time is the sum of the regularized and pure-AGF runs
($23.6+21.9$ hours). Each interval extends from a recorded resource
snapshot to the corresponding checkpoint timestamp. It includes some setup
and checkpoint saving, but excludes idle time between runs and training
beyond step 400. We use the same accounting for our SGF reproduction from
the ODE initialization. Estimates based on short timing runs exclude
loading, warmup, and checkpointing. These costs cover different training
stages and use different timing procedures; they do not measure throughput
under a common workload.

\begin{table}[!htbp]
  \centering
  \small
  \setlength{\tabcolsep}{3.5pt}
  \renewcommand{\arraystretch}{1.12}
  \caption{\textbf{Training requirements and cost accounting on eight H20 GPUs.}
  Only the listed stages are counted; upstream initialization and data
  preparation are excluded. Check marks indicate yes and dashes indicate no.}
  \label{tab:training-cost}
  \begin{tabular*}{\linewidth}{@{\extracolsep{\fill}}llrrcc@{}}
    \multicolumn{6}{@{}l}{\textit{(a) Training configuration and video-data requirements}} \\[3pt]
    \toprule
    Method & Training stage & Iterations & \shortstack{Global\\batch}
      & \shortstack{Training\\horizon} & \shortstack{Extra\\video data} \\
    \midrule
    Self-Forcing & DMD & 600 & 64 & 5 s & -- \\
    Causal Forcing & Asymmetric DMD & 750 & 64 & 5 s & -- \\
    LongLive & Long tuning & 3,000 & 64
      & \shortstack{60 s stream\\(5 s windows)} & -- \\
    SGF & ODE-init reproduction & 1,200 & 8 & 5 s & -- \\
    OPSD-V & LoRA post-training & 200 & 24 & 45 s & $\checkmark$ \\
    \textbf{AGF} & Regularized + pure AGF & 200 + 200 & 64 & 5 s & -- \\
    \bottomrule
  \end{tabular*}
  \par\vspace{8pt}
  \begin{tabular*}{\linewidth}{@{\extracolsep{\fill}}llrccr@{}}
    \multicolumn{6}{@{}l}{\textit{(b) Timing evidence and resulting budget}} \\[3pt]
    \toprule
    Method & Timing sample & \shortstack{Timed\\batch}
      & Timing basis & Extrap. & GPU-hours \\
    \midrule
    Self-Forcing & 35 / 5 & 8 & 40.5 s/iter & $\checkmark$ & 431.6 \\
    Causal Forcing & SF timing proxy & 8 & 40.5 s/iter & $\checkmark$ & 539.6 \\
    LongLive & 25 / 5 & 8 & 52.0 s/iter & $\checkmark$ & 2774.4 \\
    SGF & Full run & 8 & 19.3 h & -- & 154.8 \\
    OPSD-V & 4 / 2 & 8 & 708 s/iter & $\checkmark$ & 944.4 \\
    \textbf{AGF} & Full run & 64 & 45.5 h & -- & \textbf{364.4} \\
    \bottomrule
  \end{tabular*}
  \par\vspace{5pt}
  \begin{minipage}{\linewidth}
    \small
    Extra video data refers to the training stage listed in (a): OPSD-V uses 3,800 real long
    videos~\citep{liu2026opsdv}; LongLive uses generated text-prompt pairs,
    without video data~\citep{yang2026longlive}. Timing sample gives retained /
    excluded warmup iterations. Full run denotes a recorded operating interval;
    SF timing proxy reuses Self-Forcing's measurements, without a separate
    Causal Forcing timing run. Extrap. marks scaling to the full iteration
    count and global batch in (a).
  \end{minipage}
\end{table}

\subsection{Long-Horizon Inference Protocol}

\begin{table}[!htb]
  \centering
  \small
  \caption{Evaluation sets for each video duration. Each method produces one
  video per prompt, using the same mapping from prompt index to noise seed.}
  \label{tab:evaluation-cohorts}
  \begin{tabular}{lrrrrl}
    \toprule
    Horizon & Prompts & Frames & Seconds & Segments/video & Evaluated dimensions \\
    \midrule
    5 s & 944 & 81 & 5.0625 & 1 & 16 VBench dimensions \\
    60 s & 240 & 957 & 59.8125 & 6 (10 s) or 30 (2 s) & 16 VBench dimensions \\
    240 s & 120 & 3837 & 239.8125 & 120 & 6 dimensions \\
    \bottomrule
  \end{tabular}
\end{table}

At 60 seconds, Temporal Flickering applies an additional static-video filter and retains at
most five selected segments per prompt; consequently its valid-prompt counts
are 17/17/15/19/19/18 for
Self-Forcing/OPSD-V/SGF/LongLive/Causal Forcing/AGF rather than all 20 prompts
assigned to that dimension.
Tables~\ref{tab:all-vbench-dimensions} and
\ref{tab:all-vbench-dimensions-60} report the valid counts at each horizon.

The 5-second evaluation set contains 944 unique VBench prompts. The 60-second
set is a randomly sampled 240-prompt subset. The 240-second set contains
120 prompts randomly sampled from the 1,003-prompt
MovieGenVideoBench\_extended pool~\citep{polyak2024moviegen}.
Generation uses noise seed $100000+i$ for prompt index $i$. Across
methods, inference uses
Wan2.1-T2V-1.3B at 832$\times$480 and
16 fps, denoising nodes $[1000,750,500,250]$, guidance 3.0, timestep shift 5.0,
three frames per block, context noise 0, and local attention over 12 latent
frames: three persistent sink frames and nine rolling frames. Relative
sink positions are enabled, and no ground-truth replacement is used.

\begin{table}[!htb]
  \centering
  \small
  \caption{Model variants used in the controlled comparison. All six methods
  share the generation and evaluation protocol described above.}
  \label{tab:baseline-checkpoints}
  \begin{tabular}{p{0.22\linewidth}p{0.69\linewidth}}
    \toprule
    Method & Evaluated weights \\
    \midrule
    Self-Forcing & Self-Forcing generator checkpoint; no additional EMA selection at inference \\
    OPSD-V & Rank-256, $\alpha=256$ LoRA (dropout 0); non-EMA weights \\
    SGF & Our reproduction using the official hyperparameters and ODE initialization; generator EMA at iteration 1200 \\
    LongLive & Released generator base and generator LoRA; non-EMA weights \\
    Causal Forcing & Released final Stage-3 chunkwise generator; non-EMA weights \\
    AGF (ours) & Step-400 generator EMA after 200 regularized and 200 pure-AGF iterations \\
    \bottomrule
  \end{tabular}
\end{table}

\subsection{Metrics and Statistical Protocol}

At 60 seconds, aesthetic quality, imaging quality, human action, temporal
style, and overall consistency use nominal 10-second clips; the other
dimensions use two-second clips, except appearance style, which uses
two-second clips for 18 prompts and 10-second clips for two prompts. Each
dimension averages its valid segment scores before the normalization
below. Temporal flickering instead averages the evaluator's per-video
aggregates over the videos retained by its static-content filter. The valid
counts for each dimension appear in Table~\ref{tab:all-vbench-dimensions-60}.

To compute the reported 5- and 60-second scores, we normalize each raw
dimension as $(x-l)/(u-l)$. The VBench normalization bounds used here are: subject $[0.1462,1]$, background
$[0.2615,1]$, flickering $[0.6293,1]$, motion smoothness $[0.706,0.9975]$,
scene $[0,0.8222]$, appearance style $[0.0009,0.2855]$, and temporal style and
overall consistency $[0,0.364]$; all remaining dimensions use $[0,1]$.
Dynamic Degree has weight 0.5 in the quality mean and every other dimension has
weight 1. Quality and semantic subscores are combined as $0.8Q+0.2S$.

At 240 seconds, we define two summaries of the six custom-prompt dimensions
in VBench-Long~\citep{huang2025vbenchpp}. Q6 is the unweighted mean of subject
consistency, background consistency, motion smoothness, aesthetic quality,
imaging quality, and Dynamic Degree. Q5 averages the first five dimensions,
which are also used in LongLive's interactive evaluation~\citep{yang2026longlive}.
Whole-video Q5/Q6 average the
per-video dimension scores and then average across prompts. Neither composite
is an official VBench total. For the temporal versions, subject and background
use their local \texttt{inclip\_score}, whereas the whole-video scores also
include clip-to-clip consistency. The remaining dimensions use their local
segment score. Local Q5 therefore tracks quality within successive intervals;
it does not by itself establish identity or scene correspondence to the opening
of the video.

For Figure~\ref{fig:temporal-decay}, ten consecutive two-second positions form
each non-overlapping 20-second bin. We subtract each prompt's first-bin Q5 from
its later bins before averaging over prompts. Pointwise 95\% percentile
bootstrap intervals use 10,000 resamples of the 120 prompt indices (seed
20260907). Each resample retains complete trajectories and uses the same
indices across methods, preserving temporal and cross-method pairing.
These bands quantify prompt-level uncertainty in this evaluation setting;
they are not simultaneous confidence bands or per-video prediction intervals.
The comparison does not estimate variation across generation seeds, training
checkpoints, or evaluator versions.

\FloatBarrier
\begingroup
\setlength{\intextsep}{4pt plus 1pt minus 1pt}
\section{Complete Multi-Length Results}
\label{app:complete-benchmark}

\begin{table}[H]
  \centering
  \scriptsize
  \renewcommand{\arraystretch}{0.94}
  \caption{Complete raw dimension scores at 5 seconds. Each $n$ entry gives
  valid prompt counts in Self-Forcing/OPSD-V/SGF/LongLive/Causal Forcing/AGF
  order. Scores are percentages; the reported summary scores are computed after normalization.}
  \label{tab:all-vbench-dimensions}
  \resizebox{\textwidth}{!}{%
  \begin{tabular}{llrrrrrr}
    \toprule
    Dimension & $n$ & Self-Forcing & OPSD-V & SGF & LongLive & Causal Forcing & AGF \\
    \midrule
    Subject consistency & 72/72/72/72/72/72 & 96.32 & 96.08 & 96.73 & 97.47 & 95.68 & 96.52 \\
    Background consistency & 86/86/86/86/86/86 & 95.84 & 95.24 & 95.45 & 97.07 & 95.70 & 95.05 \\
    Temporal flickering & 75/75/75/75/75/75 & 99.00 & 98.55 & 97.97 & 99.31 & 98.81 & 98.85 \\
    Motion smoothness & 72/72/72/72/72/72 & 98.65 & 98.43 & 98.23 & 98.94 & 98.21 & 98.52 \\
    Dynamic degree & 72/72/72/72/72/72 & 72.22 & 73.61 & 73.61 & 31.94 & 84.72 & 70.83 \\
    Aesthetic quality & 93/93/93/93/93/93 & 62.29 & 62.62 & 64.69 & 63.71 & 64.65 & 63.92 \\
    Imaging quality & 93/93/93/93/93/93 & 70.16 & 70.35 & 70.74 & 69.39 & 71.27 & 70.55 \\
    Object class & 79/79/79/79/79/79 & 81.41 & 81.65 & 84.02 & 81.88 & 83.62 & 82.28 \\
    Multiple objects & 82/82/82/82/82/82 & 67.15 & 63.41 & 72.87 & 70.12 & 66.62 & 67.23 \\
    Human action & 100/100/100/100/100/100 & 82.00 & 78.00 & 77.00 & 83.00 & 80.00 & 80.00 \\
    Color & 81/79/81/80/82/82 & 92.70 & 92.04 & 92.36 & 92.68 & 89.06 & 90.90 \\
    Spatial relationship & 84/84/84/84/84/84 & 79.84 & 83.72 & 80.87 & 81.08 & 78.38 & 79.21 \\
    Scene & 86/86/86/86/86/86 & 32.63 & 36.56 & 27.91 & 27.76 & 30.31 & 30.38 \\
    Appearance style & 90/90/90/90/90/90 & 19.15 & 18.92 & 19.32 & 19.05 & 19.35 & 19.39 \\
    Temporal style & 100/100/100/100/100/100 & 23.41 & 23.30 & 23.27 & 22.87 & 24.31 & 23.29 \\
    Overall consistency & 93/93/93/93/93/93 & 24.49 & 24.26 & 24.16 & 24.15 & 24.58 & 24.34 \\
    \bottomrule
  \end{tabular}}
\end{table}

\begin{table}[!htb]
  \centering
  \scriptsize
  \renewcommand{\arraystretch}{0.94}
  \caption{Complete raw dimension scores at 60 seconds. The count and method
  order follow Table~\ref{tab:all-vbench-dimensions}.}
  \label{tab:all-vbench-dimensions-60}
  \resizebox{\textwidth}{!}{%
  \begin{tabular}{llrrrrrr}
    \toprule
    Dimension & $n$ & Self-Forcing & OPSD-V & SGF & LongLive & Causal Forcing & AGF \\
    \midrule
    Subject consistency & 20/20/20/20/20/20 & 97.51 & 97.46 & 97.83 & 98.36 & 95.07 & 97.91 \\
    Background consistency & 42/42/42/42/42/42 & 96.32 & 95.82 & 95.92 & 96.78 & 94.87 & 96.56 \\
    Temporal flickering & 17/17/15/19/19/18 & 99.06 & 98.53 & 98.54 & 99.43 & 99.02 & 99.14 \\
    Motion smoothness & 20/20/20/20/20/20 & 98.53 & 98.39 & 98.49 & 98.78 & 97.05 & 98.58 \\
    Dynamic degree & 20/20/20/20/20/20 & 61.33 & 61.67 & 57.50 & 35.83 & 87.83 & 68.50 \\
    Aesthetic quality & 20/20/20/20/20/20 & 55.08 & 57.07 & 59.92 & 58.17 & 56.18 & 58.48 \\
    Imaging quality & 20/20/20/20/20/20 & 68.06 & 69.87 & 70.80 & 68.79 & 69.38 & 71.05 \\
    Object class & 20/20/20/20/20/20 & 79.53 & 79.95 & 75.33 & 81.22 & 73.07 & 84.05 \\
    Multiple objects & 20/20/20/20/20/20 & 63.90 & 69.75 & 66.98 & 85.67 & 53.95 & 72.75 \\
    Human action & 20/20/20/20/20/20 & 77.50 & 79.17 & 76.67 & 80.83 & 76.67 & 77.50 \\
    Color & 20/20/20/20/20/20 & 88.87 & 86.31 & 80.14 & 92.51 & 83.84 & 93.77 \\
    Spatial relationship & 20/20/20/20/20/20 & 70.48 & 80.03 & 74.72 & 71.65 & 76.36 & 71.37 \\
    Scene & 42/42/42/42/42/42 & 19.44 & 21.37 & 16.59 & 29.93 & 18.87 & 24.46 \\
    Appearance style & 20/20/20/20/20/20 & 20.06 & 19.44 & 19.47 & 19.50 & 20.06 & 19.78 \\
    Temporal style & 20/20/20/20/20/20 & 20.81 & 21.12 & 19.80 & 20.96 & 20.93 & 20.72 \\
    Overall consistency & 20/20/20/20/20/20 & 22.88 & 22.50 & 21.95 & 22.29 & 21.98 & 22.79 \\
    \bottomrule
  \end{tabular}}
\end{table}

\begin{table}[!htb]
  \centering
  \scriptsize
  \renewcommand{\arraystretch}{0.94}
  \caption{All six dimensions evaluated at 240 seconds on the same 120 prompts.
  Values are percentages.}
  \label{tab:all-240s-dimensions}
  \begin{tabular}{lrrrrrr}
    \toprule
    Dimension & Self-Forcing & OPSD-V & SGF & LongLive & Causal Forcing & AGF \\
    \midrule
    Subject consistency & 96.36 & 96.29 & 97.11 & 97.08 & 94.44 & \textbf{97.32} \\
    Background consistency & 95.79 & 95.64 & 96.10 & 96.24 & 94.82 & \textbf{96.38} \\
    Motion smoothness & 97.95 & 97.91 & 98.18 & \textbf{98.60} & 95.99 & 98.31 \\
    Aesthetic quality & 54.97 & 56.60 & 58.32 & 58.29 & 55.29 & \textbf{61.20} \\
    Imaging quality & 67.26 & 67.77 & 70.63 & 67.86 & 68.41 & \textbf{70.91} \\
    Dynamic degree & 63.19 & 65.10 & 51.08 & 49.53 & \textbf{91.41} & 62.54 \\
    \midrule
    Q6 (all six dimensions) & 79.25 & 79.88 & 78.57 & 77.94 & \textbf{83.39} & 81.11 \\
    Q5 (excludes Dynamic Degree) & 82.47 & 82.84 & 84.07 & 83.62 & 81.79 & \textbf{84.82} \\
    \bottomrule
  \end{tabular}
\end{table}
\FloatBarrier
\endgroup

\begingroup
\raggedbottom
\FloatBarrier
\section{Additional Long-Horizon Examples}
\label{app:additional-cases}

\setlength{\parskip}{3pt}
\setlength{\intextsep}{2pt plus 1pt minus 1pt}
\setlength{\textfloatsep}{6pt plus 1pt minus 1pt}
\setlength{\floatsep}{6pt plus 1pt minus 1pt}
\setlength{\abovecaptionskip}{2pt}
\setlength{\belowcaptionskip}{0pt}

These 240-second rollouts show how scene composition, dynamic interactions,
and subject and background structure evolve over time. Every row shows
the same six time points: 0, 48, 96, 144, 192, and 240 seconds. Labels report
whole-video scores in percent: \emph{Dynamic Degree} and
\emph{Five-metric Quality}, the unweighted mean of subject consistency,
background consistency, motion smoothness, aesthetic quality, and imaging
quality (Q5). Original prompt excerpts appear below each panel.

\subsection{Camera Drift and Loss of Scene Structure}
\label{app:dynamic-degree}

Higher Dynamic Degree does not necessarily indicate better visual quality:
a larger motion score can accompany camera drift or structural degradation.
The following cases show AGF preserving clearer content and obtaining
higher Q5 at lower Dynamic Degree.

\begin{figure}[!t]
  \centering
  \begin{minipage}{0.97\linewidth}
    \input{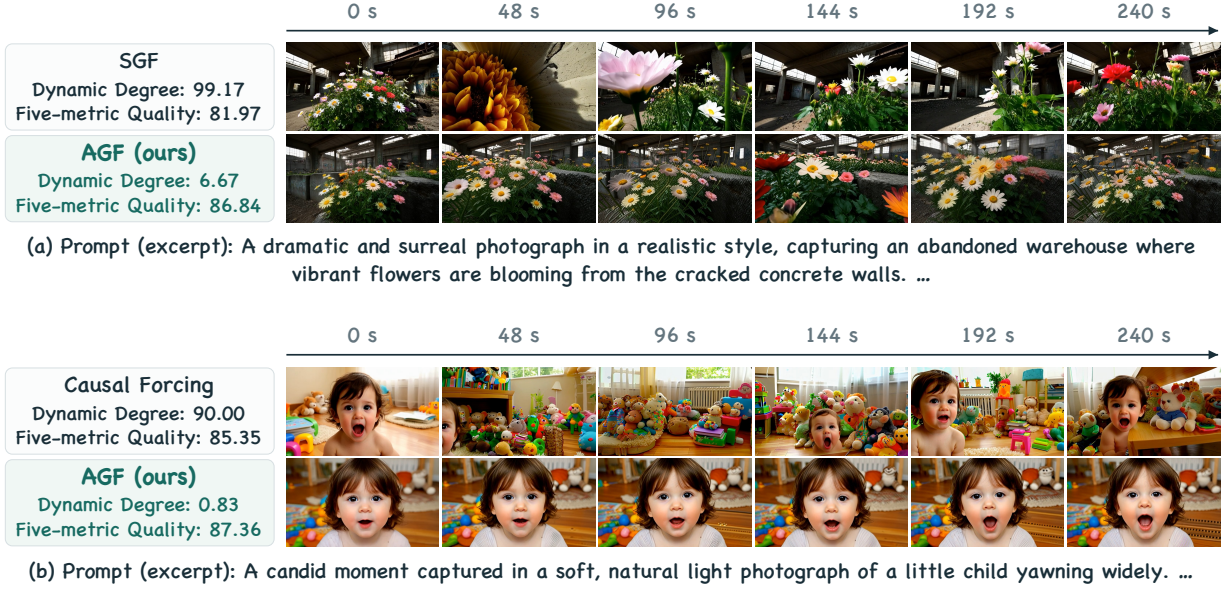}
  \end{minipage}
  \caption{\textbf{Camera drift changes the scene composition.}
  \textbf{(a) Warehouse flowers:} SGF shifts into close-ups that lose the
  industrial surroundings, while AGF keeps the flowers and warehouse
  visible together. \textbf{(b) Yawning child:} Causal Forcing moves the view
  toward the toys and partly or fully excludes the child; AGF retains a
  clear view of the child and the changing expression.}
  \label{fig:dd-camera}
\end{figure}

\begin{figure}[!t]
  \centering
  \begin{minipage}{0.97\linewidth}
    \input{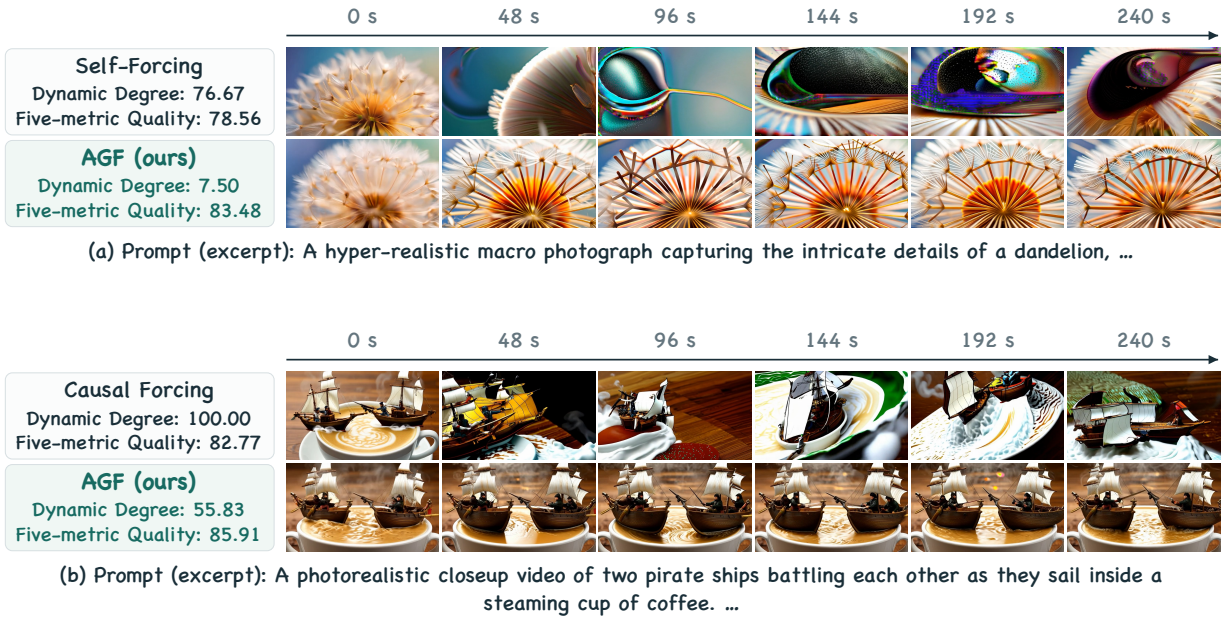}
  \end{minipage}
  \caption{\textbf{Loss of fine detail and object structure.}
  \textbf{(a) Dandelion:} distorted surfaces and colored artifacts replace
  the fine seed structure in Self-Forcing, while AGF retains the radial
  organization and visible filaments. \textbf{(b) Pirate ships:} Causal
  Forcing distorts the foam, cup, and ship placement; AGF preserves two
  recognizable ships within a clearly defined cup.}
  \label{fig:dd-corruption}
\end{figure}

In Figure~\ref{fig:dd-camera}(a), SGF alternates between warehouse views
and extreme flower close-ups, repeatedly changing how the flowers are
framed within the building. AGF keeps the flowers and industrial interior
visible together in a consistent composition. In
Figure~\ref{fig:dd-camera}(b), Causal Forcing redirects the view from the
child toward the toys: the child is partly cropped at 48 seconds and absent
at 96 seconds. AGF keeps the child's face centered as the mouth and
expression change.

Figure~\ref{fig:dd-corruption} examines fine detail and object layout.
In the dandelion sequence, Self-Forcing's later frames replace the fine
seed structure with large distorted surfaces and colored artifacts.
AGF retains the recognizable radial structure and fine filaments
through the sampled close-ups. In the coffee-cup sequence, Causal Forcing
increasingly warps the foam and cup geometry, while the ships lose their
stable placement within the cup. AGF keeps both ships, the foam, and the
cup rim recognizable as parts of the same miniature scene.

\FloatBarrier

\setlength{\parskip}{1pt}
\Needspace{90pt}
\subsection{Coherent Dynamic Scenes}
\label{app:high-dynamic-quality}

AGF can retain high visual quality while generating dynamic scenes. The
examples below pair its high Dynamic Degree with coherent object
manipulation and character movement.

\begingroup
\setlength{\intextsep}{0pt}
\setlength{\abovecaptionskip}{0pt}
\linespread{0.94}\selectfont
\begin{figure}[H]
  \centering
  \begin{minipage}{0.95\linewidth}
    \input{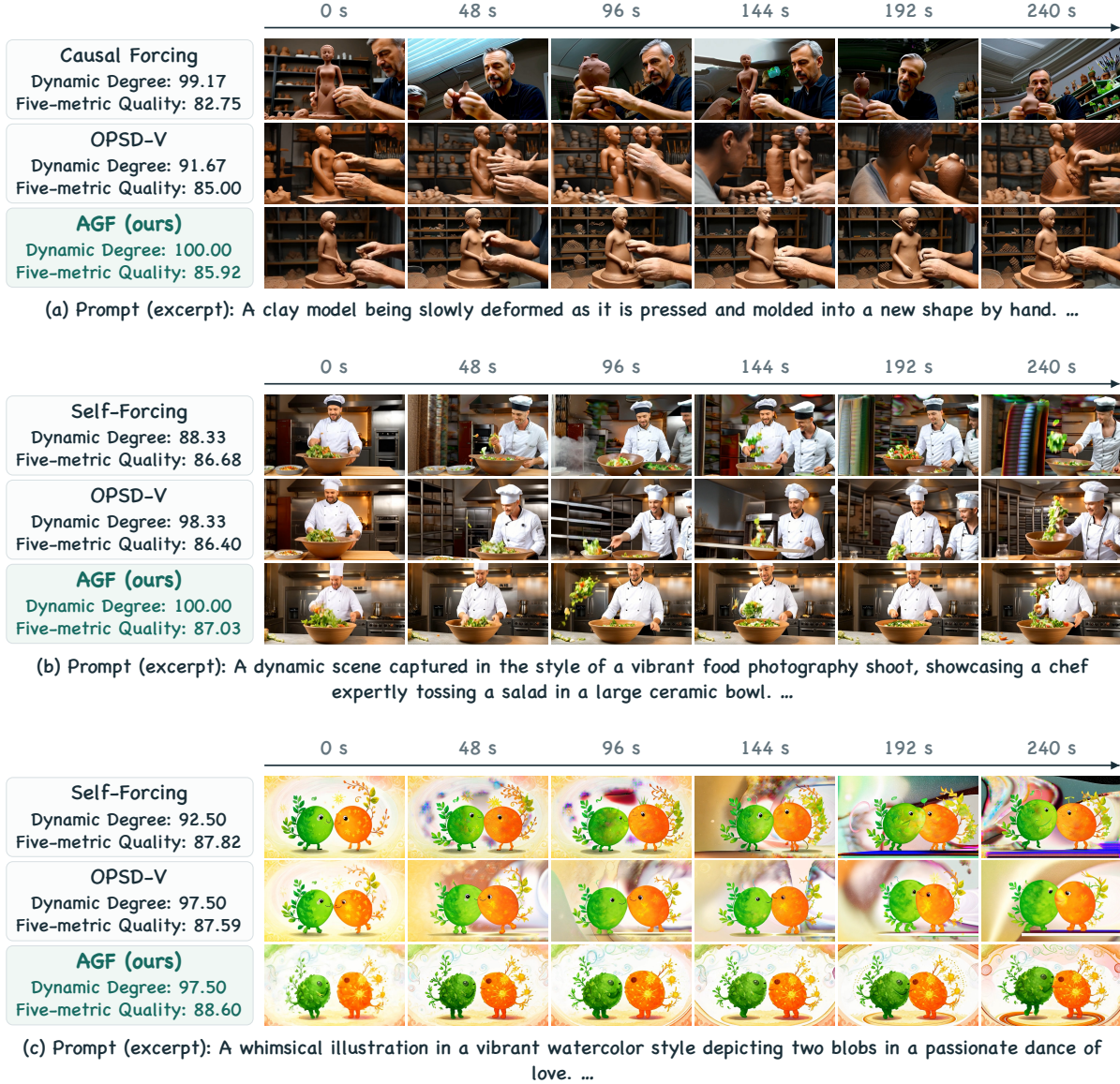}
  \end{minipage}
  \caption{\textbf{Maintaining subjects and settings during dynamic actions.}
  \textbf{(a) Sculpting:} OPSD-V introduces adjacent copies of the clay
  figure; Causal Forcing repeatedly reframes the interaction around the
  sculptor's face, taking the work surface out of view. AGF keeps one
  figure on the work surface as the hands shape it. \textbf{(b) Salad
  tossing:} Self-Forcing develops repeated chefs and background striping;
  OPSD-V also duplicates the chef. AGF retains one chef, the bowl, and the
  kitchen. \textbf{(c) Dancing characters:} the alternatives develop
  background deformations and colored bands, while AGF preserves the green
  and orange pair and their pastel setting as they move.}
  \label{fig:high-dynamic-cases}
\end{figure}
\endgroup
\setlength{\parskip}{4pt}
\setlength{\intextsep}{4pt plus 1pt minus 1pt}
\setlength{\abovecaptionskip}{4pt}

Figure~\ref{fig:high-dynamic-cases} shows hands reshaping clay, a chef
tossing salad, and two colored characters changing pose and position.
AGF preserves the objects and spatial relations involved in each action:
the hands work on one clay figure, tossed ingredients appear
above the bowl, and the two characters remain distinct in their pastel
setting. The studio, kitchen, and illustrated background remain
recognizable through the sampled views.

\FloatBarrier

\Needspace{90pt}
\subsection{Subject Duplication and Background Corruption}
\label{app:similar-quality}

Similar Q5 scores can conceal substantial differences in visual coherence.
In the following cases, other methods achieve Q5 close to or above AGF's
while introducing subject copies or background corruption. The aggregate
score does not fully reflect these visible differences.

\begin{figure}[H]
  \centering
  \begin{minipage}{0.95\linewidth}
    \input{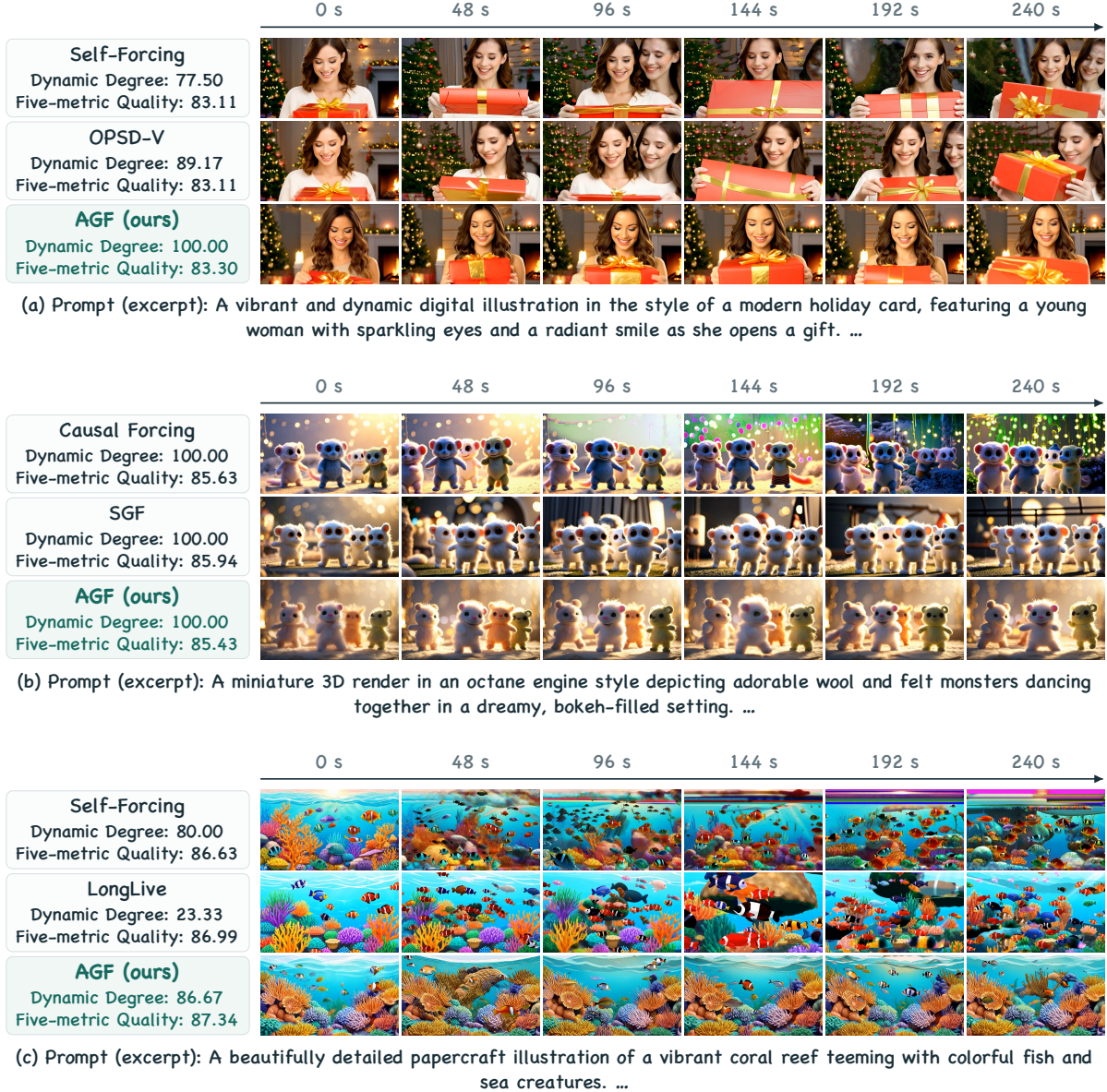}
  \end{minipage}
  \caption{\textbf{Subject copies and background artifacts in long rollouts.}
  \textbf{(a) Gift:} Self-Forcing and OPSD-V add copies of the woman and
  alter the room; AGF retains one main subject and the holiday setting.
  \textbf{(b) Felt monsters:} Causal Forcing and SGF replace the softly
  blurred background with patterned scenery or indoor structures, while
  AGF retains the soft bokeh. \textbf{(c) Papercraft reef:} Self-Forcing
  adds colored stripes and LongLive introduces oversized dark forms;
  AGF keeps the coral environment and papercraft texture recognizable.}
  \label{fig:similar-quality-cases}
\end{figure}

In Figure~\ref{fig:similar-quality-cases}(a), Self-Forcing and OPSD-V
introduce additional copies of the woman and alter the room, while AGF
retains one main subject and the holiday setting. In
Figure~\ref{fig:similar-quality-cases}(b), Causal Forcing replaces the soft
bokeh with bright patterned scenery, and SGF introduces curtains and
windows. AGF retains the felt appearance of the monsters and their softly
blurred surroundings. In Figure~\ref{fig:similar-quality-cases}(c),
Self-Forcing develops horizontal colored stripes, while LongLive introduces
oversized dark forms that interrupt the reef composition. AGF preserves a
recognizable coral environment and layered papercraft texture as the fish
change position.

\FloatBarrier

\Needspace{90pt}
\subsection{Complete Six-Method Comparisons}
\label{app:full-case-comparisons}

\begin{figure}[!htbp]
  \centering
  \input{figure/source/appendix_six_methods.tex}
  \caption{\textbf{Two additional scenes across all six methods.}
  \textbf{(a) Bookshelf mushrooms:} several alternatives develop repetitive
  geometry or striping, or shift away from the shelf. AGF preserves a clear
  view of the mushrooms on the shelf. \textbf{(b) Dragon and underworld:}
  close-ups and distorted geometry disrupt the relation between the dragon
  and people in several alternatives; AGF keeps the dragon, people, and
  cavern recognizable together.}
  \label{fig:long-horizon-cases-appendix}
\end{figure}

The full six-method comparisons in
Figure~\ref{fig:long-horizon-cases-appendix} complement the main
example of a church on the coast with two further scenes. In both, AGF combines the
highest Q5 with clearer preservation of the main subjects and their spatial
relationships.

In the bookshelf scene, close-ups, repeated structures, and colored bands obscure
the mushrooms and their surroundings in several other methods. AGF keeps the
mushrooms and their placement on the shelf recognizable. In the underworld
scene, viewpoint and geometry changes disrupt the composition around the
dragon and torch-bearing people; AGF keeps both visible within a
recognizable cavern.

\endgroup

\section{Additional Temporal Stability Results}
\label{app:temporal-results}

\subsection{Quality over Unperturbed Rollouts}
\label{app:temporal-unperturbed}

Table~\ref{tab:temporal-decay-summary} gives the absolute local Q5 scores
underlying Figure~\ref{fig:temporal-decay}. Each entry averages a consecutive
20-second interval over the same 120 prompts, with one video per prompt.
AGF stays near its initial quality throughout the four-minute rollout,
while all five baselines finish below their initial scores.

\begin{table}[H]
  \centering
  \small
  \setlength{\tabcolsep}{7pt}
  \caption{Local Q5 (\%) over unperturbed four-minute generation.
  Q5 averages the five quality dimensions and excludes Dynamic Degree.
  Bold marks the highest mean in each interval.}
  \label{tab:temporal-decay-summary}
  \begin{tabular}{lrrrrrr}
    \toprule
    Time (s) & \shortstack{Self-\\Forcing} & OPSD-V & SGF & LongLive
      & \shortstack{Causal\\Forcing} & \shortstack{AGF\\(ours)} \\
    \midrule
    0--20 & 84.19 & 83.75 & \textbf{84.85} & 84.58 & 82.73 & 84.71 \\
    20--40 & 83.50 & 83.06 & 84.39 & 83.87 & 81.68 & \textbf{84.65} \\
    40--60 & 82.77 & 82.62 & 84.07 & 83.70 & 81.35 & \textbf{84.58} \\
    60--80 & 82.45 & 82.57 & 83.95 & 83.38 & 81.29 & \textbf{84.65} \\
    80--100 & 82.33 & 82.58 & 83.85 & 83.32 & 81.45 & \textbf{84.71} \\
    100--120 & 82.18 & 82.41 & 83.61 & 83.34 & 81.29 & \textbf{84.60} \\
    120--140 & 82.23 & 82.48 & 83.77 & 83.37 & 81.37 & \textbf{84.69} \\
    140--160 & 81.73 & 82.50 & 83.78 & 83.31 & 81.29 & \textbf{84.66} \\
    160--180 & 81.83 & 82.39 & 83.96 & 83.27 & 81.07 & \textbf{84.61} \\
    180--200 & 81.76 & 82.41 & 84.03 & 83.24 & 81.05 & \textbf{84.63} \\
    200--220 & 81.56 & 82.31 & 83.97 & 83.23 & 81.38 & \textbf{84.69} \\
    220--240 & 81.58 & 82.32 & 83.89 & 83.45 & 81.27 & \textbf{84.74} \\
    \bottomrule
  \end{tabular}
\end{table}

\subsection{Stability after Temporary History Perturbations}
\label{app:temporal-perturbed}

We evaluate the same six methods on the 120 prompts with two inference-time
perturbations. Both act from 30 to 60 seconds on each newly generated latent
block before it is encoded into the history cache; saved outputs are not
edited. At 60 seconds, we stop injecting errors and continue generation
from the accumulated history until 240 seconds.

\begin{figure}[H]
  \centering
  \includegraphics[width=\linewidth]{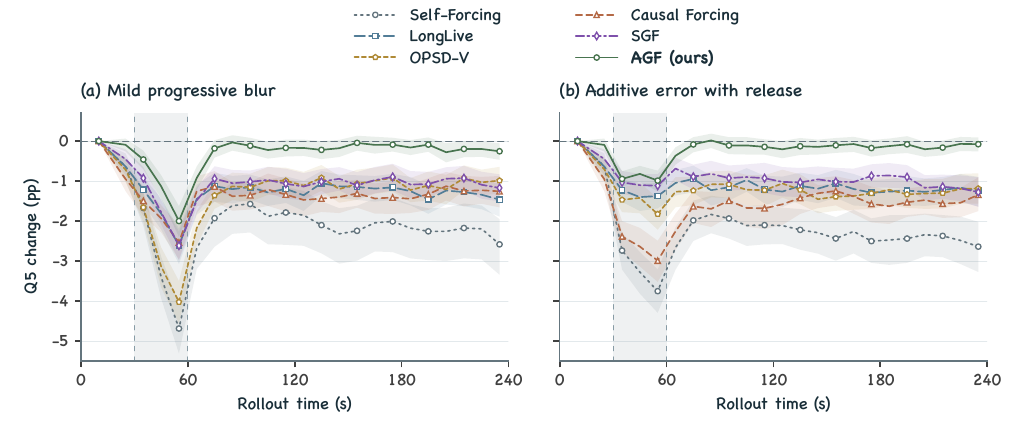}
  \caption{\textbf{Generation stability under temporary history perturbations.}
  Mean local Q5 change from each video's first 20 seconds, over 120 prompts.
  The first point summarizes 0--20 seconds; subsequent points use 10-second
  bins. Gray shading marks the 30--60-second intervention. Colored bands
  show pointwise 95\% bootstrap intervals from 2,000 prompt resamples.
  These trajectories include natural quality drift as well as the effects
  of the intervention.}
  \label{fig:inference-robustness}
\end{figure}

\paragraph{Mild progressive blur.}
For a history block $h$, we write
$\tilde h=(1-\alpha)h+\alpha\,\mathrm{AvgPool}_{5\times5}(h)$
to the cache. The spatial mean filter uses reflection padding, and
$\alpha$ increases linearly from $0.0625$ to $0.25$ across the intervened blocks.

\paragraph{Additive error with release.}
At the onset, we compute a history perturbation that increases the squared
error between the next block's prediction and a fixed denoised target from
the frozen 14B teacher. We differentiate through all four denoising steps
of that next block, sharing the perturbation across the history block's
three latent frames. The resulting gradient direction is scaled to a
nominal RMS of $0.30$ and held fixed for each method--prompt pair; the same
error field is added to every new history block during the intervention.

All six methods lose quality during the intervention and recover after
it ends (Figure~\ref{fig:inference-robustness}). AGF returns closest to its
initial level: its mean Q5 change over 90--240 seconds is $-0.17$ pp after
blur and $-0.13$ pp after additive error, compared with $-2.10$ to $-1.03$ pp
and $-2.31$ to $-1.01$ pp, respectively, for the five baselines.

\section{Additional Ablations}
\label{app:ablations}

The ablations below use the configuration in Table~\ref{tab:training-config}
with only the specified setting changed. They use Self-Forcing ODE
initialization, batch 64, five-second rollouts, and 200 regularized + 200
pure-AGF iterations. We evaluate
the generator's EMA weights at iteration 400 on the same 944/240/120 prompts
at 5/60/240 seconds;
default AGF scores reuse the main benchmark. All variants use one training
run and one video per prompt, with ordinary, unperturbed inference.

\subsection{Regularization Variants}

\paragraph{Without regularization.}
\label{app:regularizer-presence}
The variant without regularization (None in Table~\ref{tab:ablations})
uses the generator's EMA weights at iteration 400 and is evaluated
on the same 944/240/120 prompts with the common inference settings.
It obtains 81.38/77.96 Total at 5/60 seconds and 82.75 Q5 at 240 seconds.

\paragraph{Regularized steps.}
For regularization over multiple denoising steps, both history conditions use
shared noisy inputs from the clean trajectory. We average the per-step
MSEs against detached clean predictions. We weight this combined loss using
the same 5\% target ratio, based on gradient estimates at the model outputs.

\paragraph{Reference models.}
The default prediction target is the current generator's velocity prediction
at the first denoising step, conditioned on the unperturbed history. This
prediction is detached from the computation graph. The \emph{Frozen}
variant uses the same
ODE-initialized generator held fixed throughout training. The \emph{Teacher}
variant uses a frozen 14B teacher to generate a full video from the shared
initial noise with 25 sampling steps and guidance 3.5. Its output is converted
to an equivalent first-step velocity at the actual scheduler noise level;
the teacher does not condition on the student's generated history.
All three variants apply the regularizer only to the continuation under the
perturbed history. They use the same schedule for the target gradient ratio,
which reaches 5\%, as described in Appendix~\ref{app:prediction-consistency-details}.

\subsection{Alternative ODE Initialization}
\label{app:causal-init}

$\mathrm{AGF}_{\mathrm{CF}}$ starts from Causal Forcing's chunkwise Stage 2 ODE checkpoint. Given the different starting point and optimization characteristics of this initialization, we slightly adjust the training schedule and train for 1,400 iterations in total. The regularization strength ramps from zero at iteration 20 to a 3\% output-gradient ratio at iteration 80, holds through iteration 400, and is disabled for iterations 401--1,400. EMA starts at iteration 100 with decay 0.99 and is evaluated at iteration 1,400. Other settings follow Table~\ref{tab:training-config}.

Compared with the released Stage 3 Causal Forcing model, all five Q5
dimensions increase (Table~\ref{tab:causal-init-details}). Lower Dynamic
Degree makes Q6 fall despite higher Q5, reinforcing the need to report
visual quality separately from the extent of motion.

\begin{table}[H]
  \centering
  \small
  \setlength{\abovecaptionskip}{0pt}
  \setlength{\belowcaptionskip}{4pt}
  \caption{\textbf{Alternative initialization at 240 seconds (\%).}}
  \label{tab:causal-init-details}
  \setlength{\tabcolsep}{2pt}
  \renewcommand{\arraystretch}{1.05}
  \begin{tabular*}{\linewidth}{@{\extracolsep{\fill}}lrrrrrrrr@{}}
    \toprule
    Model & \shortstack{Subject\\consistency}
      & \shortstack{Background\\consistency}
      & \shortstack{Motion\\smoothness}
      & \shortstack{Aesthetic\\quality}
      & \shortstack{Imaging\\quality}
      & \shortstack{Dynamic\\Degree}  & Q5 \\
    \midrule
    CF & 94.44 & 94.82 & 95.99 & 55.29 & 68.41 & 91.41  & 81.79 \\
    $\mathrm{AGF}_{\mathrm{CF}}$ & \textbf{97.37} & \textbf{96.44}
      & \textbf{98.33} & \textbf{59.87} & \textbf{71.01}
      & 79.95 & \textbf{84.60} \\
    \bottomrule
  \end{tabular*}
\end{table}

\Needspace{90pt}
\section{Human Evaluation Details}
\label{app:visual-ranking}

The study comprises 240 ranking tasks completed by 12 participants,
covering 120 prompts. Each task presents six four-minute videos generated
from the same prompt, with method identities hidden. Participants rank
all six candidates in order of preference, assigning each rank from
1 to 6 exactly once. Each participant completes 20 tasks.

\begin{figure}[H]
  \centering
  \includegraphics[width=0.9\linewidth]{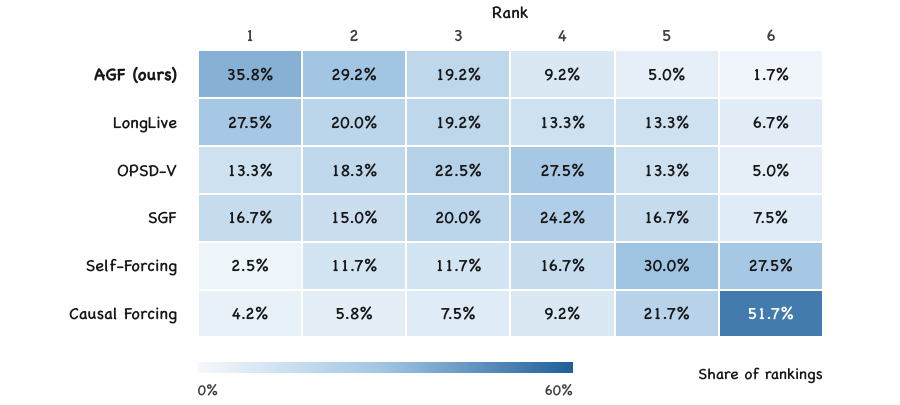}
  \caption{\textbf{Distribution of human-assigned ranks.} Each cell shows
  the percentage of 240 ranking tasks assigning that rank to the method.
  Rows follow the mean-rank ordering in Figure~\ref{fig:ranking-mean};
  darker cells indicate a larger share. Percentages are rounded to one
  decimal place.}
  \label{fig:ranking-distribution}
\end{figure}

For each method, we compute its mean rank as the sum of its assigned
ranks divided by 240. Figure~\ref{fig:ranking-distribution} shows the
percentage of assignments at each rank, using the same denominator.
This distribution complements the mean by showing how often each method
appears near the top or bottom of the ordering.

AGF leads the human evaluation with the best mean rank (2.23), the largest
share of first-place rankings (35.8\%), and a top-three placement in
84.2\% of ranking tasks. This preference trend is broadly consistent with
the four-minute quality evaluation in Table~\ref{tab:long-results}.
Together, these results show that AGF's advantage in automatic quality
metrics is also reflected in human judgments of four-minute videos.

\end{document}